\documentclass[10pt,twocolumn,letterpaper]{article}

\usepackage{iccv}      % To produce the REVIEW version
\usepackage{pifont}
\usepackage{multirow}
\usepackage{wrapfig}
\usepackage{caption}
\usepackage{bbm}
\usepackage[dvipsnames]{xcolor}
\usepackage[accsupp]{axessibility}

\definecolor{iccvblue}{rgb}{0.21,0.49,0.74}
\usepackage[pagebackref,breaklinks,colorlinks,allcolors=iccvblue]{hyperref}

\def\paperID{14020} % *** Enter the Paper ID here
\def\confName{ICCV}
\def\confYear{2025}

\title{ On the Generalization of Representation Uncertainty in Earth Observation}

\author{
Spyros Kondylatos$^{1,2,3}$\thanks{Equal contribution}\quad
Nikolaos Ioannis Bountos$^{1,2,4}$\footnotemark[1]\quad
Dimitrios Michail$^4$\\
Xiao Xiang Zhu$^{5,6}$\quad
Gustau Camps-Valls$^3$\quad
Ioannis Papoutsis$^{1,2,7}$\\
$^1$National Observatory of Athens\enspace
$^2$National Technical University of Athens\\
$^3$University of Valencia\enspace
$^4$Harokopio University of Athens\enspace
$^5$Technical University of Munich\\
$^6$Munich Center for Machine Learning\enspace
$^7$Archimedes/Athena RC\\%, Athena Research Center\\
{\tt\small \{skondylatos,bountos\}@noa.gr, michail@hua.gr, xiaoxiang.zhu@tum.de,}\\
{\tt\small gustau.camps@uv.es, ipapoutsis@mail.ntua.gr}
}

\begin{document}
\maketitle
\begin{abstract}

Recent advances in Computer Vision have introduced the concept of pretrained representation uncertainty, enabling zero-shot uncertainty estimation. 
This holds significant potential for Earth Observation (EO), where trustworthiness is critical, yet the complexity of EO data poses challenges to uncertainty-aware methods.
In this work, we investigate the generalization of representation uncertainty in EO, considering the domain's unique semantic characteristics. 
We pretrain uncertainties on large EO datasets and propose an evaluation framework to assess their zero-shot performance in multi-label classification and segmentation EO tasks. 
Our findings reveal that, unlike uncertainties pretrained on natural images, EO-pretraining exhibits strong generalization across unseen EO domains, geographic locations, and target granularities,
while maintaining sensitivity to variations in ground sampling distance.
We demonstrate the practical utility of pretrained uncertainties showcasing their alignment with task-specific uncertainties in downstream tasks, their sensitivity to real-world EO image noise, and their ability to generate spatial uncertainty estimates out-of-the-box.
Initiating the discussion on representation uncertainty in EO, our study provides insights into its strengths and limitations, paving the way for future research in the field.
Code and weights are available at: \url{https://github.com/Orion-AI-Lab/EOUncertaintyGeneralization}.

\end{abstract}    
\section{Introduction}
\label{sec:intro}

\begin{figure}
    \centering
    \includegraphics[width=\linewidth]{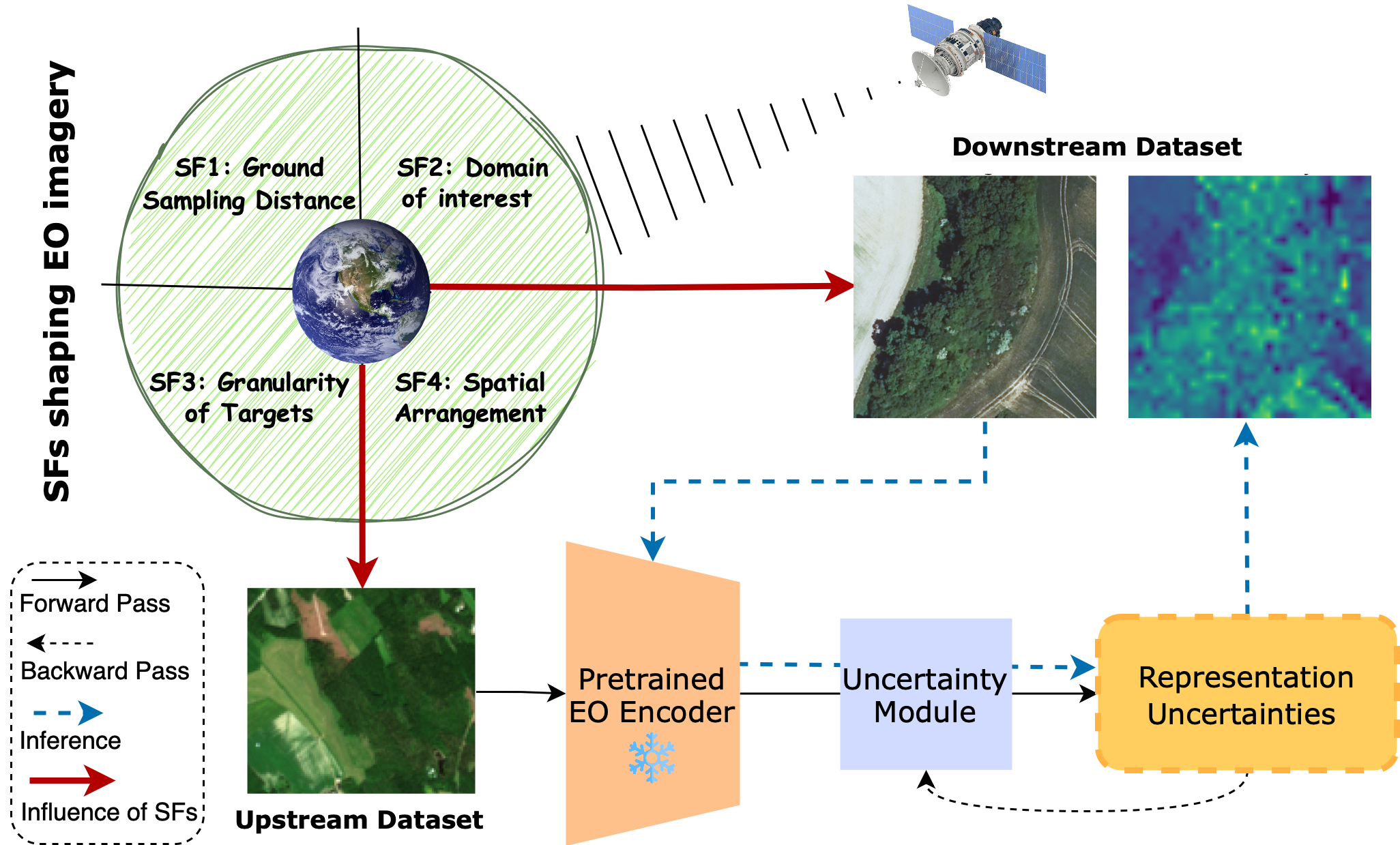}
    \caption{Overview of our investigation framework. We pretrain uncertainties on upstream frozen Earth Observation (EO) representations, enabling zero-shot uncertainty estimation and spatial uncertainty estimates for EO downstream applications. We evaluate the generalization capacity of uncertainties, considering the key Semantic Factors (SFs) shaping EO imagery.}
    \label{fig:investigation}
\end{figure}

The criticality of Earth Observation (EO) applications \cite{yang2013role, nakalembe2020urgent, dl_earth_sciences_camps-valls} demands trustworthy Deep Learning (DL) models \cite{gawlikowski_survey_2022}.
Uncertainty estimation can enhance models' reliability by providing confidence estimates for their predictions.
However, standard uncertainty-aware methods introduce increased modeling complexity, high computational overhead, and greater inference demands \cite{gal_uncertainty_nodate}. 
To overcome these limitations, recent advances in Computer Vision (CV) have introduced pretrained uncertainties \cite{kirchhof2024pretrained, kirchhof2023url}.
These uncertainties are learned in the representation space of DL models trained on large-scale upstream datasets, enabling zero-shot uncertainty transfer to downstream tasks.
Similar to representation learning \cite{kirillov_segment_2023, devlin2019bert}, this allows extracting representation uncertainties that support task-agnostic, label-free, and generalizable uncertainty estimation.

EO applications can benefit from these advances; yet, the complexity and high dimensionality of EO data hinder large-scale modeling. 
The EO community has made progress in addressing these challenges via representation learning on large EO data archives \cite{cong2022satmae,bountos2023fomo,xiong2024neural}. 
Motivated by these efforts, we explore the potential of large-scale representation uncertainty learning in EO.

We build on the framework of Kirchhof \etal \cite{kirchhof2024pretrained}, which uses a loss prediction objective \cite{yoo_learning_2019} as a proxy to aleatoric uncertainty estimation.
Uncertainties are learned by an uncertainty module built on top of embedding representations extracted from large-scale CV models and assessed on single-label tasks.
Adapting to EO, we pretrain uncertainties on large EO datasets and investigate their zero-shot generalization in various EO datasets and tasks. 
We identify key Semantic Factors (SFs), defining the distinct semantic characteristics of EO imagery (see \cref{fig:investigation} and \cref{sub:semantic_factors}), and assess their impact on the generalization of representation uncertainty.
Furthermore, we introduce a novel set of metrics for evaluating uncertainties in the representation space beyond the single-label setting, including multi-label classification and semantic segmentation tasks for the first time. 
We reveal the practical utility of zero-shot uncertainty estimations in three ways: i) by assessing their alignment with task-specific uncertainties in downstream applications, ii) by measuring their effectiveness in identifying noisy data, and iii) by providing spatial uncertainty estimations. 
This study reveals many crucial insights.
The most important can be summarized as follows:

\begin{itemize} 
    \item Major domain gap degrades uncertainty generalization in EO as models pretrained on natural images fail to produce reliable estimations, contrary to EO pretraining. This is attributed exclusively to the uncertainty module, as embedding representations transfer similarly in both cases.
    \item EO-pretrained uncertainty generalizes well to unseen geographic locations, EO tasks, and target granularities.
    \item Uncertainty generalization is highly sensitive to Ground Sampling Distance (GSD), being optimal when the spatial resolutions of pretraining and inference datasets are close.
    \item Pretrained uncertainties are reliable for zero-shot uncertainty estimation in downstream applications, consistently aligning with task-specific uncertainties.
\end{itemize}

\section{Related Work}
\label{sec:related_work}

\textbf{Large-scale pretraining in EO:} 
Public data archives like Sentinel and Landsat enable large-scale EO pretraining. 
Many studies \cite{wang2023ssl4eo,cong2022satmae} adapted Masked Autoencoders (MAE) \cite{he2022masked} to EO, but lacked diversity in GSD and spectral bands. 
Scale-MAE \cite{reed2023scale} addressed the GSD limitation, while CROMA \cite{fuller2023croma} combined contrastive learning with a reconstruction loss fusing aligned Sentinel-$1$ (S$1$) and Sentinel-$2$ (S$2$) data. 
Guo \etal \cite{guo2024skysense} used contrastive learning and cross-modal alignment for WorldView RGB, S$1$ and S$2$ data. 
Recent methods used MAE-based pretraining to build foundation models flexible in terms of spectral bands and GSD \eg, FoMo-Net \cite{bountos2023fomo} and DOFA \cite{xiong2024neural}. 
In this work, we extend beyond large-scale representation learning to explore large-scale representation uncertainty learning in EO.

\noindent\textbf{Uncertainty estimation in the representation space:} 
Recent advances in uncertainty estimation have shifted focus from classifier predictions to the representation space.
Feed-forward methods integrate uncertainty estimation directly into the model's forward pass \cite{collier_correlated_2021, collier_massively_2023}.
Alternatively, probabilistic embeddings encode representations as distributions rather than single points \cite{nakamura_representation_2023, kirchhof_probabilistic_2023, kim_probabilistic_2023}, allowing variance to serve as uncertainty measurement.
Recently, a dedicated benchmark has been introduced to evaluate transferable uncertainty estimates in representation learning \cite{kirchhof2023url}.
Other methods frame uncertainty estimation as a regression task, predicting the training loss as a proxy for uncertainty \cite{yoo_learning_2019, laves_well-calibrated_2020, lahlou_deup_2023}. 
Kirchhof \etal \cite{kirchhof2024pretrained} demonstrated the effectiveness of this approach in large-scale settings.
In this work, we build upon this method, extending it to the EO domain.

\noindent\textbf{Uncertainty estimation in EO:} Despite its critical importance, uncertainty estimation in EO remains relatively underexplored.
Gaussian Processes \cite{rasmussen_gaussian_2006} and Deep Gaussian Processes \cite{damianou_deep_2013} have been applied to EO tasks \cite{camps-valls_survey_2016, camps-valls_perspective_2019} like biophysical parameter retrieval \cite{mateo-sanchis_gap_2018, svendsen_deep_2020} and image classification \cite{morales-alvarez_remote_2018}, but their limited scalability hinders their use in large-scale EO applications. 
Other works have explored Bayesian Neural Networks for uncertainty estimation in EO \cite{martinez-ferrer_quantifying_2022, asadi_evaluation_2020, kampffmeyer_semantic_2016, joshaghani_bayesian_2023}. 
We argue that the unique characteristics and complexity of EO data present significant challenges to uncertainty-aware methods. 
Thus, providing reliable EO-pretrained uncertainties could offer substantial benefits, simplifying uncertainty estimation in this critical domain.

\label{sec:related}

\section{Investigation Framework}

This section outlines our investigation framework, detailing the method (\cref{sub:methodology}), defining the SFs that shape EO data (\cref{sub:semantic_factors}), and describing the datasets (\cref{sec:datasets}) and evaluation methods (\cref{sec:evaluation}) used for assessing uncertainty.

\begin{table*}
    \centering
    \resizebox{\textwidth}{!}{%
    \begin{tabular}{ccccccccccc}
         Dataset & Input Modality & Sensor & DL Task & \# Classes & EO Task & Spatial Res. & Pretraining & Inference & Image Size & Coverage\\
         \toprule
         ImageNet & RGB & Optical &Classification & $21,841$ & Optical Images & - & \textcolor{Green}{\ding{52}} & \textcolor{red}{\ding{56}} & $224 \times 224$ & -\\
         \midrule
         BigEarthNet & MS/SAR & S$1$, S$2$ & Multi-label classification & 19 & LULC classification & $10$m & \textcolor{Green}{\ding{52}} & \textcolor{Green}{\ding{52}} & $120 \times 120$ & Europe\\
         \midrule
         BigEarthNet-5 & MS/SAR & S$1$, S$2$ & Multi-label classification & 5 & LULC classification & $10$m & \textcolor{Green}{\ding{52}} & \textcolor{Red}{\ding{56}} & $120 \times 120$ & Europe\\
         \midrule
         MLRSNet & RGB & Multi-sensor & Multi-label classification& $60$ & Semantic Scene Understanding & $\approx$$10$ - $0.1$m & \textcolor{Red}{\ding{56}} & \textcolor{Green}{\ding{52}} & $256 \times 256$ & Global\\
         \midrule
         Woody & RGB & Aerial & Image Segmentation & $4$ &Tree-species detection  & $50$cm & \textcolor{Red}{\ding{56}} & \textcolor{Green}{\ding{52}} & $224 \times 224$ & Chile\\
         \midrule
         Waititu & RGB & Aerial &Image Segmentation & $3$ &Invasion tree-species detection & $50$cm & \textcolor{Red}{\ding{56}} & \textcolor{Green}{\ding{52}} & $224 \times 224$ & New Zealand\\
         \midrule
         Flair & RGB/NIR/DEM & Aerial &Image Segmentation & $19$ &LULC classification & $20$cm & \textcolor{Green}{\ding{52}} & \textcolor{Green}{\ding{52}} & $512 \times 512$ & France\\
         \midrule
         MARIDA & MS & S$2$ &Image Segmentation & $12$ &Marine Debris Detection & $10$m & \textcolor{Red}{\ding{56}} & \textcolor{Green}{\ding{52}} & $224 \times 224$ & Global\\
         \bottomrule
    \end{tabular}%
        }
    \caption{Overview of examined datasets and tasks. % used in this study.
    S$1$ and S$2$ stand for Sentinel $1$ and $2$, MS for Multispectral, SAR for Synthetic Aperture Radar, and NIR for Near Infrared. 
    Multi-sensor stands for satellite and aerial data with varying GSDs, DEM for Digital Elevation Model, and LULC for Land Use Land Cover.
    Our study focuses on the RGB bands, but can be seamlessly expanded to other bands and modalities.}
    \label{tab:datasets}
\end{table*}

\subsection{Methodology}
\label{sub:methodology}
Kirchhof \etal \cite{kirchhof2024pretrained} introduced an efficient approach for pretraining visual uncertainties that generalize zero-shot. 
Their method introduces an uncertainty module, implemented as a multi-layer perceptron on top of pre-trained frozen representations. 
This module is trained to predict the upstream task loss as a proxy for aleatoric uncertainty. 
For each input $x$, the model estimates a scalar uncertainty value $u(x)\in\mathbb{R}$, indicating whether the representation $e(x)$ should be trusted or not.
To prevent loss-scaling issues in downstream tasks, the authors use a ranking-based objective.
This method yielded strong results across various CV tasks. 
In this work, we examine its potential for EO applications.
Further details are available in Supplementary Material (SM) \ref{app:A}.

Building on this method, we investigate the applicability of pretrained uncertainties in the EO domain.
We train supervised classification models on large-scale EO datasets and freeze their representations. 
The uncertainty module $u$ is then trained on top to predict the ranking-based loss as a measure of uncertainty.
The generalization of the representation uncertainties is evaluated across various inference datasets through two complementary strategies: i) directly in the representation space, using a novel set of metrics, and ii) assessing their ability to provide reliable uncertainty estimation for downstream applications.
This is done by training supervised classification models on downstream tasks and measuring the alignment between their losses and the zero-shot uncertainty estimates.

\subsection{Semantic factors shaping EO Imagery} 
\label{sub:semantic_factors}

EO imagery exhibits unique characteristics that differentiate it from natural optical images. It encompasses multiple scales and dimensions, exhibiting variations in object scale, resolution, acquisition modes, and environmental conditions \cite{rolf2024mission}. 
Unlike conventional imagery, it lacks a clearly defined background, as all scene elements hold relevance.
To characterize the universal modality-invariant concepts of EO imagery, we define four SFs (visualized in SM \ref{sec:vis_sf}):
% \begin{itemize}

\noindent \textbf{SF1: Ground Sampling Distance}: GSD defines the spatial resolution of an image, representing the ground area covered by each pixel. 
GSD influences the level of spatial detail, affecting object detectability, textures, and patterns.

\noindent \textbf{SF2: Domain of interest}: EO imagery is shaped by geographic, temporal, and thematic/environmental factors. 
For instance, marine environments exhibit distinct patterns and objects compared to terrestrial scenes. 
Factors such as topography, climatic conditions, and seasonality introduce spatio-temporal variability in EO data representations.

\noindent \textbf{SF3: Granularity of Targets}: The required semantic detail varies across EO tasks, ranging from fine-grained knowledge of small-scale objects ($<1$m) \eg tree species classification, to high-level patterns ($>1$km) \eg forest detection.

\noindent \textbf{SF4: Spatial arrangement}: 
Object positioning reveals insights into the processes shaping a scene.
For instance, in deforestation monitoring, the spatial distribution of cleared land patches can indicate if deforestation is planned (\eg, agricultural expansion) or irregular (\eg, illegal logging). 

In this study, we analyze the impact of variations in these SFs on aleatoric uncertainty generalization in EO.

\subsection{Datasets}
\label{sec:datasets}
\Cref{tab:datasets} summarizes the datasets used in this study.
BigEarthNetV$2$ \cite{clasen2024reben}, which we will refer to as BigEarthNet, and MLRSNet \cite{qi2020mlrsnet} are large-scale multi-label classification datasets. 
BigEarthNet addresses a Land Use and Land Cover (LULC) classification task in Europe, using S$2$ imagery at $10$m GSD, while MLRSNet targets scene understanding with high-resolution imagery of varying GSDs ($10$m to $10$cm), with global spatial coverage. 
BigEarthNet-$5$ is a simplified version of BigEarthNet, where the $19$ original classes are merged into $5$ broad categories \ie, forest, agriculture, sea, urban, and wetlands.
Woody \cite{kattenborn2019uav} and Waititu  \cite{kattenborn2020convolutional} are high-resolution aerial datasets ($50$cm per pixel) for tree species segmentation in Chile and New Zealand respectively.
Flair \cite{ign2022flair1, garioud2023flair} is also a high-resolution ($20$cm per pixel) aerial dataset solving a LULC semantic segmentation task in France.
MARIDA \cite{kikaki2022marida} is a global dataset for marine debris detection in coastal areas and river mouths based on S$2$ at $10$m GSD.
ImageNet, with over $21$k natural image classes, has proven effective for transfer learning in both EO and CV \cite{huh2016makes,ridnik2021imagenet}.
The set of examined datasets emphasizes the diversity in ML tasks, concepts, and spatial resolutions.
In particular, the pretraining datasets \ie BigEarthNet and Flair exhibit significant variations in spatial coverage (continental vs. country-scale), GSDs ($10$m vs. $20$cm per pixel), and target granularity (fine-grained vs. high-level LULC). 

\subsection{Evaluation of representation uncertainties} 
\label{sec:evaluation}
Assessing the quality of uncertainties is challenging, as no definitive gold standard exists, even in standard prediction tasks. 
However, a fundamental principle for reliable uncertainty estimation states:
\textit{Predictions with lower uncertainty should be accurate, while inaccurate predictions should exhibit higher uncertainty} \cite{mukhoti_evaluating_2019}.
This principle naturally extends to the representation space: 
\textit{Representations with lower uncertainty should be accurate, while inaccurate representations should exhibit higher uncertainty.}
Yet, defining accuracy in the representation space presents challenges, which we address in \cref{sec:metrics}.

An advantage of large-scale pretraining is that it can learn representations that improve generalization and performance in downstream tasks.
Following the same principle, pretrained uncertainties should also transfer effectively, enabling reliable zero-shot uncertainty estimation in predictive applications. 
Thus, beyond assessing uncertainties within the representation space, we also evaluate how well they align with uncertainties derived from task-specific models in downstream applications (See \cref{sub:downstream_evaluation}).

\subsubsection{Evaluation in the representation space}
\label{sec:metrics}

This section defines accuracy in the representation space and metrics for the evaluation of representation uncertainty.

\paragraph{Accuracy in the representation space:} 
Recall@$1$ \cite{pmlr-v119-roth20a} has been effectively used as a measure of accuracy in the representation space \cite{kirchhof2023url}, estimating it as the semantic similarity between representations.
It is a binary metric that determines whether a sample representation $x$ and its nearest neighbor belong to the same class.
However, Recall@$1$ is limited to single-label tasks.
Since most EO applications extend beyond this setting, we introduce \textit{Label Agreement@$1$ (LA@$1$)}, a set of metrics generalizing Recall@$1$ for multi-label classification and semantic segmentation. 
We define multiple versions of LA@$1$, each reflecting a different aspect of semantic similarity. 
An overview of the metrics is provided in the next paragraphs, with detailed formulations presented in SM \ref{app:Β}.

\noindent\textbf{LA@$1$ for multi-label classification:} 
The definition of LA@$1$ naturally extends from single-label to multi-label classification. 
We introduce three metrics: (1) \textit{One-LA@$1$}, assessing whether $x$ and $x^*$ share at least one common class, (2) \textit{All-LA@$1$}, measuring if $x^*$ contains all classes of $x$, and (3) \textit{\%-LA@$1$}, quantifying the proportion of classes of $x^*$ that match those of $x$, providing a balance between the lenient One-LA@$1$, and the stricter All-LA@$1$.

\noindent\textbf{LA@$1$ for semantic segmentation:} 
In semantic segmentation, the concept of \textit{"sharing the same class"} for an image is ambiguous, as each pixel belongs to a dedicated class.
We distinguish between two types of semantic similarity: i) spatial similarity, measuring whether the same spatial locations/pixel(s) share the same classes, and ii) contextual similarity, measuring whether images share the same classes, regardless of their spatial positions. 
A representation may exhibit high similarity in one type and low in the other.
For example, two images with reversed spatial layouts may appear spatially dissimilar yet have high contextual similarity. 

We introduce four complementary metrics to capture both types of semantic similarity:
(1) \textit{All-LA@$1$}, a high-level contextual metric assessing if $x$ and $x^*$ contain the same set of classes, (2) \textit{Patches$_{p}$-LA@$1$} measuring spatial patch-level similarity by dividing $x$ and $x^*$ into $p \times p$ patches and assessing the similarity of the majority classes across corresponding patches, (3) \textit{Probability Distance of LA@$1$ (PD-LA@$1$)} measuring contextual similarity by computing the distance of class distributions of $x$ and $x^*$, irrespective of their spatial positions, and (4) \textit{Patches$_{p}$-Probability Distance of LA@$1$ (Patches$_{p}$-PD-LA@$1$)}, combining both types of semantic similarity by calculating the average distance of class distributions of $x$ and $x^*$ across patches.

\paragraph{Uncertainty evaluation in the representation space: } 
In single-label tasks, representation uncertainty has been assessed using R-AUROC \cite{kirchhof2023url}, measuring the AUROC between predicted uncertainties $u(x)$ and Recall@$1$.
Since Recall@$1$ reflects accuracy in the representation space, effective uncertainty estimates should align with it (\eg, low Recall@1 should correspond to high uncertainty), leading to a high R-AUROC.
However, extending this to LA-AUROC, is not straightforward as $\%$-LA@$1$, PD-LA@$1$, and Patches$_{p}$-PD-LA@$1$ produce continuous values within $\left[0, 1\right]$, making traditional AUROC, which assumes binary targets, unsuitable. 
For this, we use the Coefficient of Predictive Ability (CPA) \cite{gneiting_receiver_2021}, a generalization of AUROC designed for continuous, linearly ordered targets (See SM \ref{app:cpa}).
In the binary case, CPA reduces to AUROC. 
From here on, all applications of CPA to LA@$1$, will be referred to as LA-CPA and follow the LA@1 naming convention \eg $\%$-LA-CPA refers to CPA applied to $\%$-LA@$1$. 
Each metric provides a distinct perspective on representation uncertainty.
For instance, high Patches$_{p}$-LA-CPA but low All-LA-CPA suggests that: \textit{``spatial relationships should be trusted in low-uncertainty representations, whereas they are less reliable when uncertainty is high. 
However, in both cases, it remains unclear whether contextual similarity is preserved.``}
An optimal model should excel across all metrics, ensuring both spatial and contextual reliability. 
However, metrics can also be used separately depending on the task.

\subsubsection{Evaluating uncertainties in downstream tasks}
\label{sub:downstream_evaluation}
To assess the practical utility of pretrained uncertainties, we compare their alignment with task-specific uncertainty proxies, here, the loss of supervised models trained on downstream tasks. 
For this, we use the Discard Test \cite{haynes2023creating} (SM \ref{app:discard_test}), where the most uncertain samples of the downstream task are iteratively removed based on the zero-shot pretrained uncertainties. 
The average downstream loss is computed on the remaining samples in each iteration.
A reliable model should demonstrate a decrease in error as more uncertain samples are discarded, indicating that high zero-shot uncertainty corresponds to higher downstream errors.

\begin{figure*}
    \centering
    \includegraphics[width=0.95\linewidth]{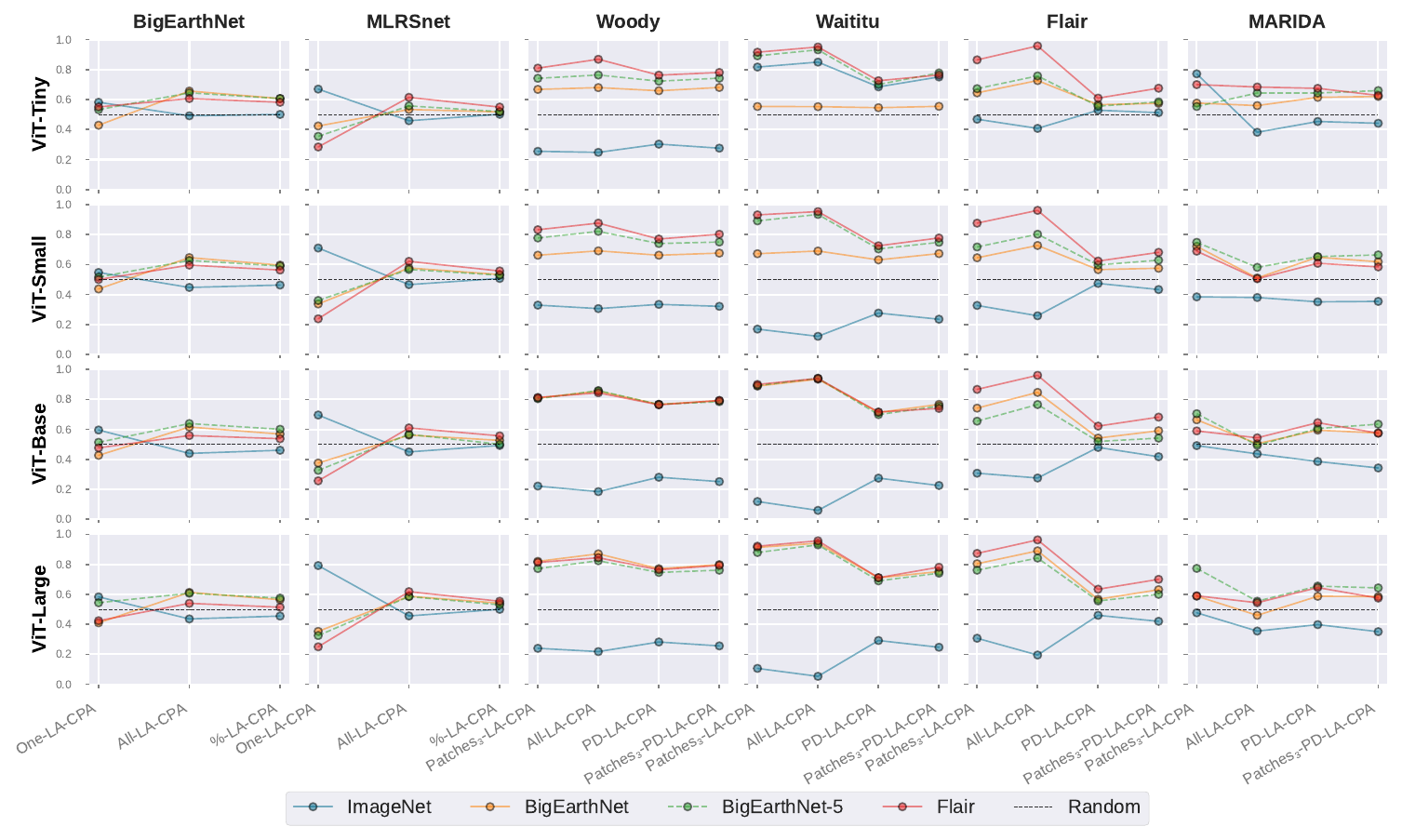}
    \caption{Performance evaluation of pretrained uncertainties across all ViT variants and inference datasets using all LA-CPA metrics, with higher values reflecting better performance. Colors indicate different pretraining datasets. EO-pretrained models consistently outperform ImageNet-pretrained models, except in One-LA-CPA, where LA@$1$ is close to $1$ (See SM \ref{app:recall@1}), reducing the interpretability of CPA.}
    \label{fig:results}
\end{figure*}

\section{Experiments \& Discussion}
\label{sec:4}

We pretrain four ViT \cite{dosovitskiy2020image} variants: ViT-Tiny, Small, Base, and Large, %all with patch size at $16\times16$ pixels,
on two well-curated EO datasets, BigEarthNet and Flair, one optical \ie, ImageNet, as well as BigEarthNet-5.  
For ImageNet models, we utilize publicly available weights by \cite{kirchhof2024pretrained}. 
To ensure consistency across pretraining setups, we convert the Flair upstream task from semantic segmentation to multi-label classification, using the unique classes in the segmentation mask as ground truth.
We conduct our experiments on RGB bands to ensure a fair comparison with ImageNet-pretrained models and to assess uncertainty generalization across datasets with varying spectral characteristics, beyond RGB (See SM \ref{app:semantic_factors} \& \ref{app:unc_module} for experiment details). 
Notably, our framework can be applied to any spectral bands and modalities; we include an additional study on BigEarthNet for Synthetic Aperture Radar (SAR) and Multispectral (MS) data in SM \ref{app:ms_sar}. 

\subsection{Generalization of pretrained uncertainties}
\label{sub:results}
In standard transfer learning, pretrained models generalize better when the source and target domains are close.
Intuitively, this principle should extend to representation uncertainty. 
To test this, we examine whether EO-pretrained uncertainties generalize better than those pretrained on generic optical data to unseen EO tasks.

\begin{figure}
    \centering
    \includegraphics[width=0.95\linewidth]{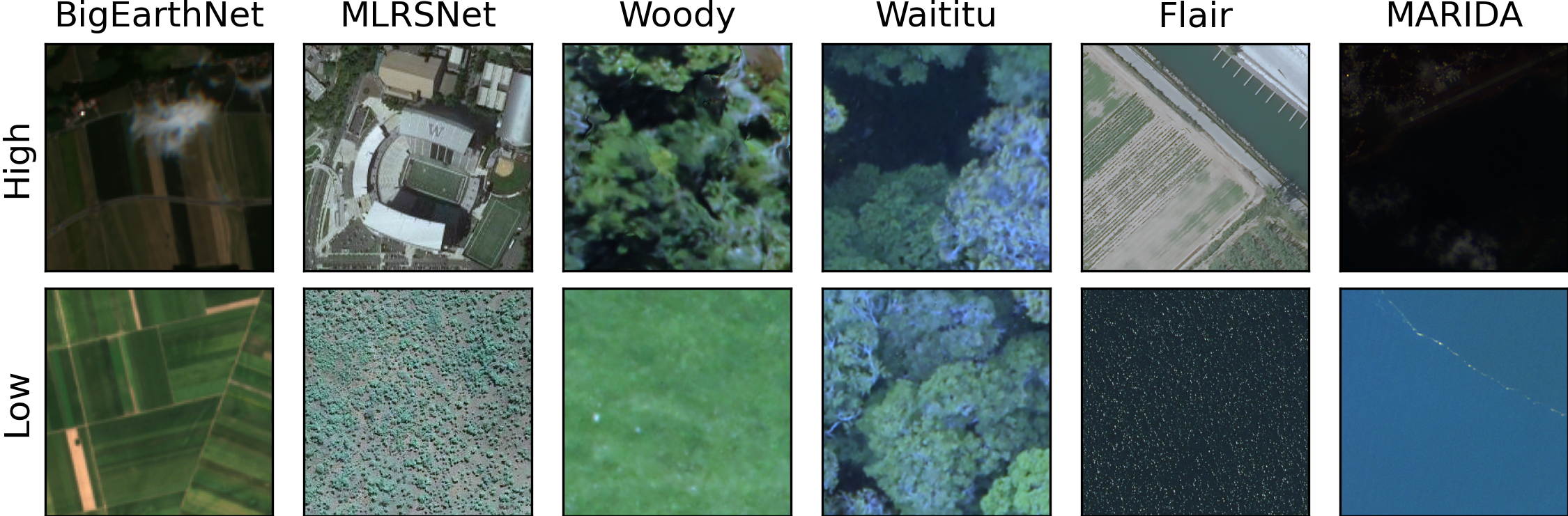}
    \caption{Samples with high/low zero-shot uncertainty, estimated by a ViT-Large pretrained on Flair.}
    \label{fig:uncertain_samples}
\end{figure}

\paragraph{EO pretraining yields reliable uncertainty estimates:}
\Cref{fig:results} shows that pretraining on EO domain produces uncertainty estimates that generalize well across our diverse benchmark, consistently outperforming both the random baseline (R-CPA = 0.5), which assigns uncertainties arbitrarily, and optical pretraining. 
Performance gains are observed in almost all evaluation metrics and models, indicating the capacity of the models to capture multiple dimensions of uncertainty. 
The only exception is One-LA-CPA, which is lower than both the random baseline and ImageNet pretraining.
This is likely due to One-LA@$1$ being close to $1$ in these cases  (See SM \ref{app:recall@1}), leading to a very small number of failed samples and a highly imbalanced One-LA@$1$ distribution. CPA and AUROC are less interpretable in these scenarios. 
A small decline is observed in contextual distribution-based metrics PD-LA-CPA \& Patches$_{p}$-PD-LA-CPA, indicating a slight misalignment between the uncertainties and the class distribution proximity.
\Cref{fig:uncertain_samples} and SM \ref{app:unc_samples} present certain/uncertain samples for all datasets, as estimated by a ViT-Large pretrained on Flair, offering a qualitative evaluation of uncertainties. 
Simpler images exhibit lower uncertainty, while more complex scenes with multiple objects or cloud cover yield higher uncertainty.

\paragraph{Pretraining on optical data does not generalize to EO:} 
Models pretrained on ImageNet consistently fail to provide reliable uncertainties in EO datasets, underperforming both EO pretraining and the random guess baseline. 
\textbf{This highlights the strong impact of the domain gap on the generalization of uncertainties in EO.}
Yet, ImageNet encoders generate robust representations with comparable LA@$1$ to EO-pretrained models (See SM \ref{app:recall@1}). 
Thus, \textbf{the improved uncertainty estimates of EO pretrained models stem from the domain-specific uncertainty module, not from the domain-specific encoder}. 
Since the uncertainty module is trained to predict the pretraining task's loss, the quality of the estimated uncertainty is related to the affinity between the upstream and downstream domain. 
Here, the upstream task refers to real-world objects \eg \textit{cats}, while the downstream focuses on Earth's surface objects.

\paragraph{EO pretrained uncertainties are robust to domain drift:}
While the impact of the domain gap between optical and EO data on uncertainty generalization has been established, its effect within the EO domain (SF2) needs further study, given its known effect in standard transfer learning \cite{tuia2016domain}.
To test this, we evaluate generalization to MARIDA, a dataset focusing on unseen marine concepts rather than land.
As shown in \cref{fig:results}, EO-pretrained models on land datasets exhibit comparable LA-CPA values on MARIDA and downstream datasets with smaller domain gaps (\eg, pretraining on BigEarthNet and inference on MLRSnet). 
ViT-Tiny excels in all metrics, while larger models exhibit similar performance except for the ALL-CPA criterion.
We hypothesize that this is due to large sea areas with small, similar objects scattered across images, making ALL-CPA sensitive to their presence or absence. 
The generally robust performance in MARIDA shows that \textbf{EO uncertainty pretraining can generalize to unseen EO tasks}, indicating resilience to SF2.
Moreover, \textbf{pretrained uncertainties generalize to new geographic regions, despite limited spatial coverage during pretraining}---BigEarthNet spans Europe, while Flair France. 
Notably, pretraining uncertainties on Flair yields the highest LA-CPA across all metrics for Woody (Chile) and Waititu (New Zealand), despite previously unseen environmental conditions and land properties.

\begin{figure*}
    \centering
    \includegraphics[width=1.0\linewidth]{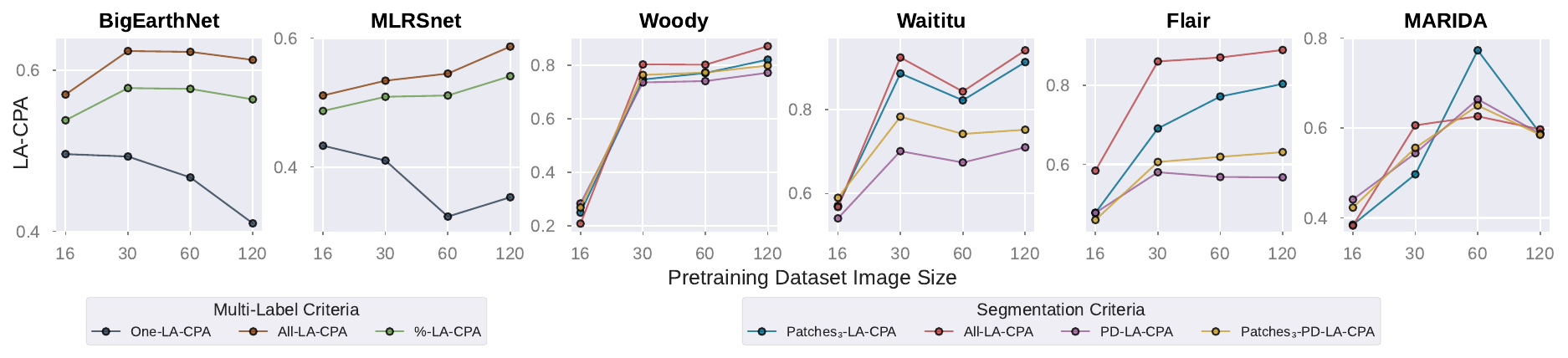}
    \caption{Evaluation of uncertainties across inference datasets when pretraining ViT-Large on BigEarthNet with varying image sizes. Colors indicate different metrics. This plot underscores the impact of GSD (SF1) on uncertainty performance.}
    \label{fig:resolution1}
\end{figure*}

\begin{figure*}
    \centering
    \includegraphics[width=1.0\linewidth]{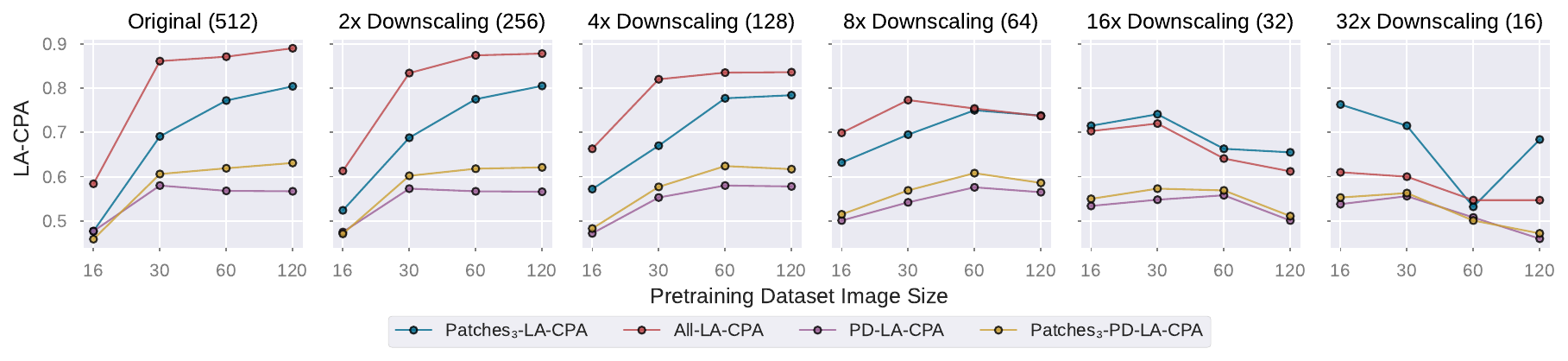}
    \caption{Evaluation of uncertainties on Flair with decreasing image size during inference, when pretraining ViT-Large on BigEarthNet with varying image sizes. As inference resolution decreases lower-resolution pretrained models gradually outperform higher-resolution ones, emphasizing the impact of GSD on uncertainty generalization.}
    \label{fig:resolution2}
\end{figure*}

\subsection{Investigating the impact of concept granularity}
\label{sub:concept}
Concept granularity depends on GSD (SF1) and target resolution \ie, the level of detail dictated by the task (SF3).
In this section, we explore the impact of these factors on the generalization of pretrained uncertainties. 
\paragraph{Uncertainty generalization is robust to changes in pretraining target granularity:}
\label{para:targets}
Intuitively, models trained on finer labels should yield more reliable uncertainties as they encode richer semantic information. 
To test this, we pretrain uncertainties on BigEarthNet-$5$, a dataset with reduced target granularity compared to BigEarthNet.  
Surprisingly, smaller models (ViT-Tiny and ViT-Small) trained on BigEarthNet-$5$ consistently outperform those trained on BigEarthNet across most datasets and metrics (See \cref{fig:results}), while larger models (ViT-Base and ViT-Large) maintain comparable performance.
We hypothesize that coarser labels emphasize fundamental semantic patterns rather than subtle, task-specific details, which are enough to preserve generalization in the examined scales. 
\textbf{Our findings show that SF3 does not significantly affect uncertainty generalization, indicating robustness to granularity reduction}, though further analysis is required for larger scales.

Reducing target granularity indirectly affects the spatial arrangement (SF4) of objects, as previously co-existing classes are merged \eg, arable land \& permanent crops. 
Thus, the similar performance of the two datasets also suggests robustness in spatial rearrangements. 

\paragraph{GSD affects generalization:}
\label{para:resolution}
\Cref{fig:results} shows that EO-pretrained uncertainties can generalize even to datasets with higher GSD than the pretraining dataset (\eg, pretraining on BigEarthNet and inference on Woody \& Waititu). 
However, models pretrained on Flair, a high-resolution dataset, yield more reliable uncertainty estimates in high-resolution inference tasks. 
This raises the question: \textit{How does GSD (SF1) affects the generalization of pretrained uncertainties?}

To investigate this, we pretrain separate ViT-Large models on BigEarthNet, each with a different input resolution: $120\times120$ to $60\times60$, $30\times30$, and $16\times16$ pixels.
We conduct inference on all datasets, keeping their original resolution intact. 
As shown in \cref{fig:resolution1}, pretraining on the original resolution yields the best performance for MLRSnet, Woody, Waititu, and Flair, indicating once more that higher-resolution pretraining produces more reliable representations for high-resolution datasets.
For the lower-resolution datasets, BigEarthNet and MARIDA, training at $60\times60$ and $30\times30$ pixels suffices.
Extreme downscaling at $16\times16$ resolution degrades performance across all datasets, except for the One-LA-CPA criterion in multi-label tasks. 

Having examined the impact of GSD during pretraining, we now evaluate the effect of downstream GSD on uncertainty generalization. 
For this, we systematically downscale Flair's resolution during inference ($2\times$, $4\times$, $8\times$, $16\times$, and $32\times$), while maintaining the same pretraining scheme as earlier (ViT-Large pretrained on BigEarthNet at varying GSDs). 
As shown in \cref{fig:resolution2}, performance initially improves with an increased pretraining GSD, but this improvement gradually diminishes as downscaling increases.
At $8\times$ downscaling, it plateaus, and beyond this point, it declines as pretraining GSD increases. 
This suggests that as Flair’s GSD decreases, the uncertainties learned from lower-resolution data become more relevant for the downstream task, likely due to the emphasis on more abstract semantic concepts. 
Based on these findings, we posit that \textbf{aligning the resolution between pretraining and inference is essential for optimal generalization}.

\begin{figure}
    \centering
    \includegraphics[width=\linewidth]{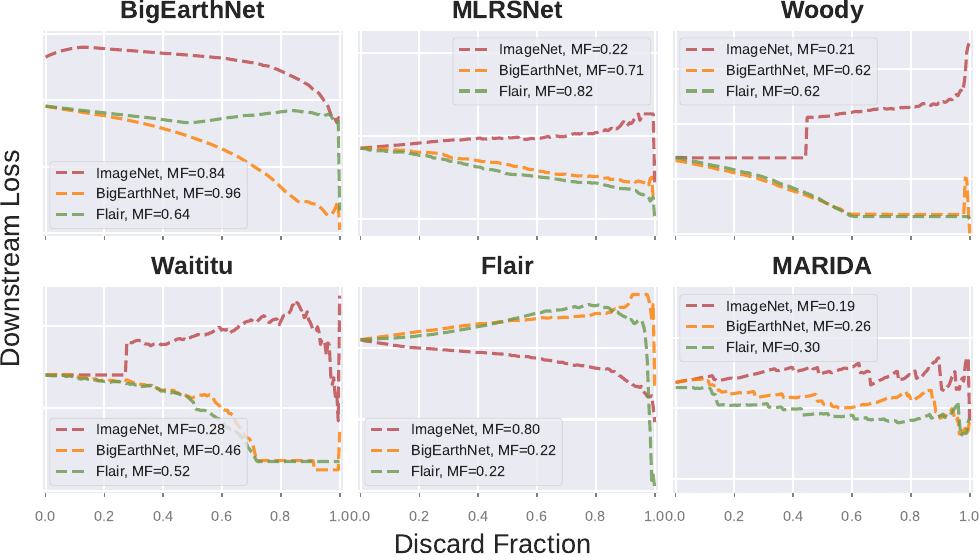}
    \caption{Discard tests for ViT-Large pretrained models across all datasets. 
    A reliable model should exhibit decreasing downstream loss as more uncertain samples---identified via zero-shot uncertainties---are discarded.
    MF (Monotonicity Fraction) indicates how often the loss decreases when samples are removed.}
    \label{fig:discard-test}
\end{figure}
\subsection{Pretrained \& task-learned uncertainties align}
\label{sub:downstream}
 
A key question remains: \textit{Are pretrained uncertainties useful in real-world applications?}
To answer this, we employ the Discard Test to assess the alignment between pretrained uncertainties and the losses of task-specific supervised models (See SM \ref{app:pre_trained}). 
\Cref{fig:discard-test} presents the Discard Test plots for all downstream datasets, where the most uncertain samples---identified via zero-shot uncertainties---are iteratively removed, and the average loss of the supervised model is computed on the remaining samples.  
For segmentation, the per-sample loss is calculated as the mean loss across all pixels of the sample.
Remarkably, increasing the discard fraction (\ie, removing more uncertain samples) generally leads to decreased task loss when pretraining on EO datasets.
This indicates that the most uncertain samples correlate with higher downstream loss values. 

\begin{figure}
    \centering
    \includegraphics[width=0.98\linewidth]{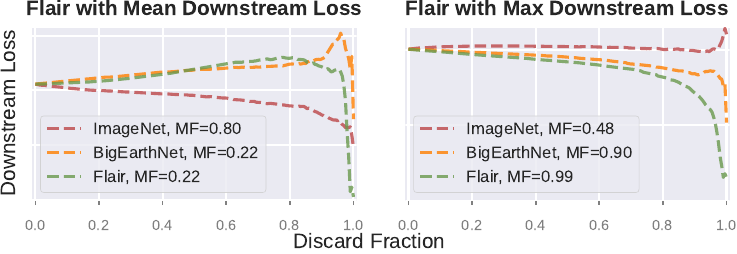}
    \caption{Discard test for ViT-Large on Flair. Uncertainties align with max pixel loss (right) but not mean pixel loss (left), capturing the most uncertain regions but not overall uncertainty.}
    \label{fig:discard_flair}
\end{figure}

Flair is the only exception to this conclusion, with all EO pretrained models diverging from this trend, and ImageNet exhibiting the best performance. 
We hypothesize that averaging per-pixel losses fails to capture sample uncertainty when only specific parts of the image are uncertain.
To investigate further, we redefine image loss as the maximum pixel loss rather than the mean, treating an image as uncertain as its most uncertain pixel. 
\Cref{fig:discard_flair} shows that EO models exhibit a nearly ideal trend in this case, while ImageNet models fail to align with the downstream loss, suggesting that highly uncertain regions can significantly influence the overall uncertainty estimate. 
These findings lead to two key conclusions:
First, \textbf{pretrained uncertainties are reliable for zero-shot uncertainty estimation in downstream applications}.
Second, aggregating an image's uncertainty into a single value may be unreliable, highlighting the need for localized uncertainty estimates in EO.

\subsection{Need for localized uncertainty estimates}
\label{sub:per_patch_unc}

\begin{figure}
    \centering
    \begin{subfigure}{0.49\linewidth}
        \centering
        \includegraphics[width=\linewidth]{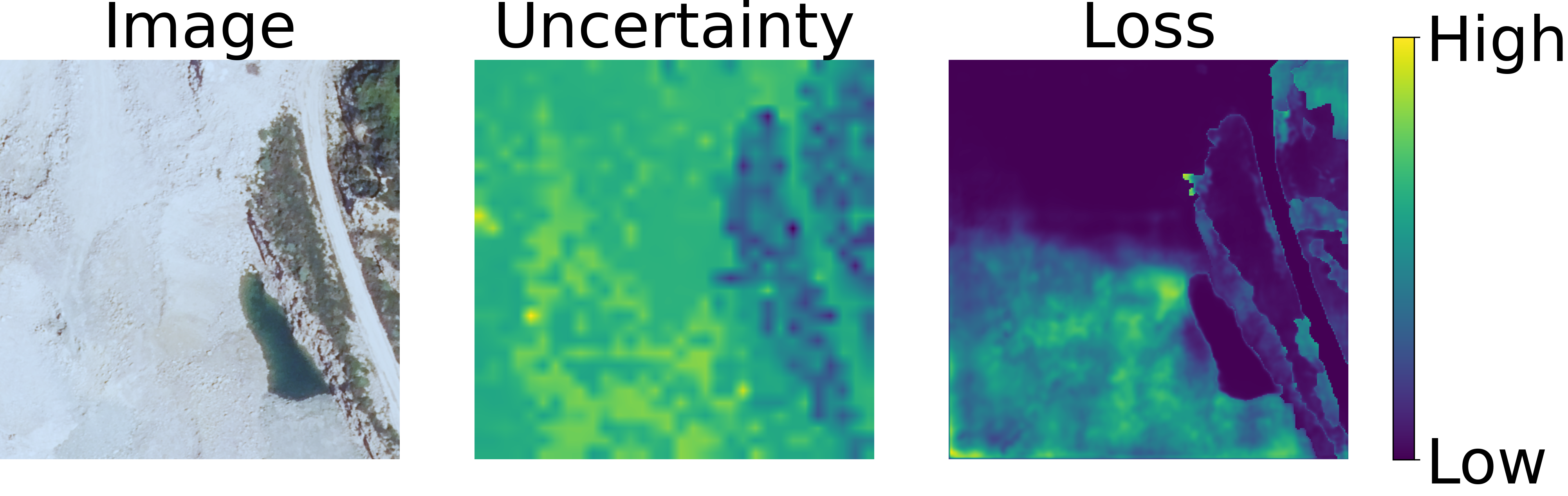}
        \caption{Flair}
        \label{fig:uncertainty}
    \end{subfigure}
    \hfill
    \begin{subfigure}{0.49\linewidth}
        \centering
        \includegraphics[width=\linewidth]{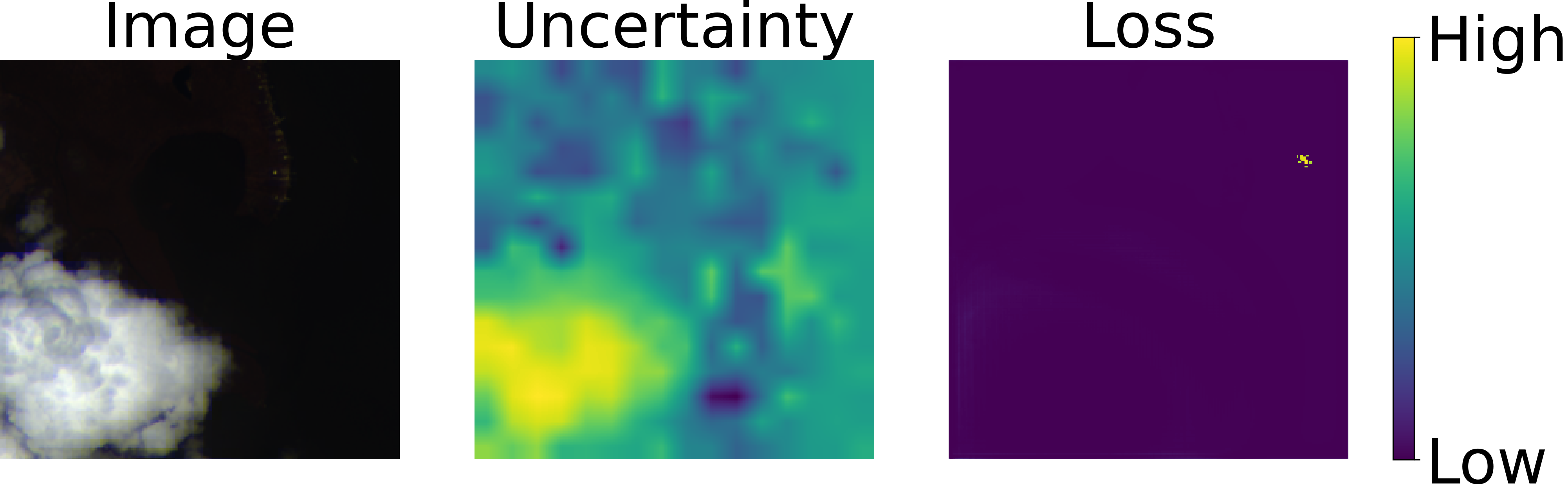}
        \caption{MARIDA}
        \label{fig:per_pixel_loss}
    \end{subfigure}
    \caption{Uncertainty of Flair \& MARIDA produced by BigEarthNet pretrained ViT-Large vis-à-vis downstream pixel loss.}
    \label{fig:per_pixel}
\end{figure}

Using a single uncertainty value per image proved to be overly restrictive in EO. 
We take advantage of the Transformer's sequence-to-sequence nature to compute per-patch uncertainties, enabling localized uncertainty estimation. 
In particular, we extract the embedded token sequence from the ViT and feed each token independently into the trained uncertainty module. 
Each token represents a spatial area of $p\times p$ pixels, where $p$ is the ViT's patch size, in this case $16$. 
\Cref{fig:per_pixel} shows representative examples for Flair and MARIDA, comparing localized uncertainties from the ViT-Large model pretrained on BigEarthNet, with the respective per-pixel loss of a task-specific U-Net \cite{ronneberger2015u}. 
The results indicate that the models produce reasonable per-patch uncertainty estimates, \eg, higher uncertainty over snow-covered areas in Flair and cloud-covered regions in MARIDA.
However, since the model is not explicitly trained to capture spatial uncertainty, the estimates do not always align with per-pixel loss. 
These initial results highlight a promising direction for future work. See SM \ref{app:localized} for more examples.

\subsection{Evidence of aleatoric uncertainty}
\label{sub:noisy}
\begin{figure}
    \centering
    \includegraphics[width=0.95\linewidth]{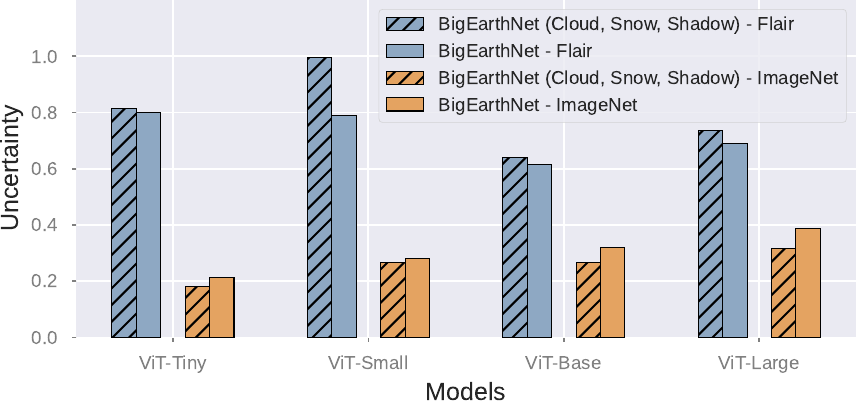}
    \caption{Comparison of uncertainties between the noise-free and noisy parts of BigEarthNet uncertainty comparison, for models pretrained on Flair (blue) and ImageNet (orange).}
    \label{fig:noisy_data_eval}
\end{figure}

%In this section,
To validate the aleatoric nature of the uncertainties, we examine their sensitivity in real-world EO noise. %the sensitivity of uncertainties in real-world EO noise.
For this, we use samples discarded from BigEarthNet due to extensive cloud, cloud shadow, and snow coverage, factors that induce aleatoric uncertainty in the images.
In \cref{fig:noisy_data_eval}, we use models pretrained on Flair and ImageNet to compare uncertainty estimates for the noisy and noise-free BigEarthNet test set.
In contrast to ImageNet pretraining, Flair-pretrained models assign higher uncertainties to the noisy dataset across all ViT variants.
This indicates that EO-pretrained models capture aleatoric uncertainty.
The uncertainty magnitudes vary across models as they depend on the pretrained classification loss. 
Additionally, similarly to \cite{kirchhof2024pretrained}, we confirm the lack of epistemic uncertainty by showing that uncertainties between the in-distribution upstream and out-of-distribution downstream data are similarly distributed (SM \ref{sec:signs_aleatoric}).

\section{Conclusion}
\label{sec:conclusion}

In this work, we investigated the generalization of representation uncertainty in EO.
For this, we introduced a concrete evaluation framework, with promise for application to broader vision fields.
Doing so, we uncovered several insights: (1) domain gap significantly impacts uncertainty generalization in EO; (2) EO-pretrained uncertainties generalize well across unseen EO domains and varying targets; (3) GSD highly affects generalization, while aligning the upstream and downstream resolution improves it; (4) pretrained uncertainties are reliable for zero-shot uncertainty estimation in real-world EO applications. 
Moreover, we extended the current method by introducing spatial uncertainties.
A limitation of this approach is that these uncertainties are estimated on representations learned on image-level tasks.
Learning uncertainty directly on dense prediction tasks is a promising future direction. % for future work.
% We extended the current method by introducing spatial uncertainties, highlighting this as a promising research direction. 
% A limitation of our approach, however, is that these spatial uncertainties are estimated on representations learned on image-level tasks rather than dense prediction ones.
Looking ahead, we envision a universal encoder capable of estimating uncertainty across a wide range of data types and modalities, paving the way for EO foundation uncertainty models.
\section*{Acknowledgments}
\label{sec:acknowledgment}
This work has received funding from the projects ThinkingEarth (grant agreement No 101130544) and MeDiTwin (grant agreement No 101159723) of the European Union’s Horizon Europe research and innovation programme.
{
    \small
    \bibliographystyle{ieeenat_fullname}
    \bibliography{main}
}

\clearpage

\maketitlesupplementary
\renewcommand{\thesection}{\Alph{section}}
\setcounter{page}{1}
\setcounter{section}{0}

\section{Optimization of pretrained uncertainties}
\label{app:A}
Loss prediction provides a general approach for uncertainty estimation, as any task’s level of wrongness can be defined by its loss $\mathcal{L}_{task}$.
In loss prediction tasks, an uncertainty module $u$ is added after the representation layer of a standard supervised encoder, predicting the corresponding loss for each sample.
Specifically, $u$ is implemented as a small MLP head, on top of the model's representation $e(x)$ and is trained using an $\mathcal{L}_2$ loss between $u$ and the task loss ${L}_{task}$.
The main supervised task is learned together with the uncertainty module, using the objective:

\begin{equation*}
    \mathcal{L} = \mathcal{L}_{task}(y, f(x)) + (u(e(x)) - \mathcal{L}_{task}(y, f(x)))^2
\end{equation*}

Kirchhof \etal \cite{kirchhof2024pretrained}, adapted the loss prediction approach to develop pretrained uncertainties, introducing some modifications. 
Particularly, they propose two methods to perform the prediction task.
First, they apply a stop-gradient mechanism before the uncertainty module to enable parallel training ensuring that its gradients do not affect the supervised classifier.
Second, they pretrain the large-scale supervised classifier and extract its representations.
The uncertainty module is trained on top of these frozen representations, enhancing computational efficiency. Since the task loss depends only on the representations, they can be cached once, accelerating training.
In this study, we adopt the second approach.

Moreover, instead of relying on the $\mathcal{L}_2$ which is inherently tied to the scale of the supervised task loss, a ranking-based objective is introduced. 
This objective ensures that uncertainty values remain consistent across different tasks and loss scales.
The ranking-based loss is defined as: 
\begin{multline*}
    \mathcal{L} = \max (0, \mathbbm{1}_{\mathcal{L}}(u(e(x_1)) - u(e(x_2)) + m)),\\
    \text{s.t. } \mathbbm{1}_{\mathcal{L}} = \left\{
\begin{array}{ll}
      +1, \text{ if } \mathcal{L}_{task}^{det}(y_1, f(x_1)) >  \mathcal{L}_{task}^{det}(y_2, f(x_2)) \\
      - 1, else
\end{array} 
\right. 
\end{multline*}

$\mathbbm{1}_{\mathcal{L}}$ is the indicator function, taking a value of $+1$ if the examined sample has higher task loss than a randomly selected sample in the batch, and $-1$ otherwise. 
During training the uncertainty values are adjusted accordingly, ensuring that the sample with higher task loss receives a higher uncertainty value, enforced by a margin of $m$.
Following \cite{kirchhof2024pretrained}, we set $m=0.1$.

\section{Visual examples of Semantic Factors}
\label{sec:vis_sf}

\begin{figure}[htbp]
    \centering
    \begin{subfigure}{0.49\linewidth}
        \centering
        \includegraphics[width=0.78\linewidth]{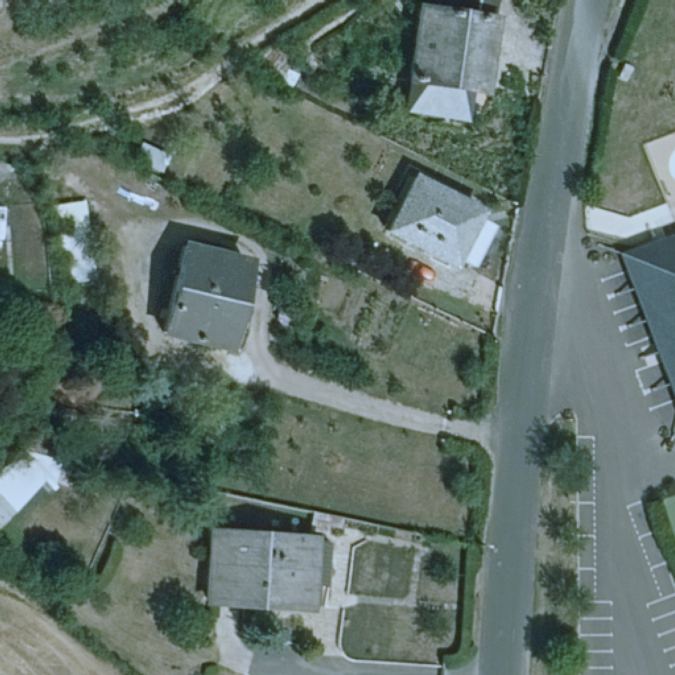}
        \caption{GSD: 20cm per pixel}
        \label{fig:sf1_example_a}
    \end{subfigure}
    \hfill
    \begin{subfigure}{0.49\linewidth}
        \centering
        \includegraphics[width=0.78\linewidth]{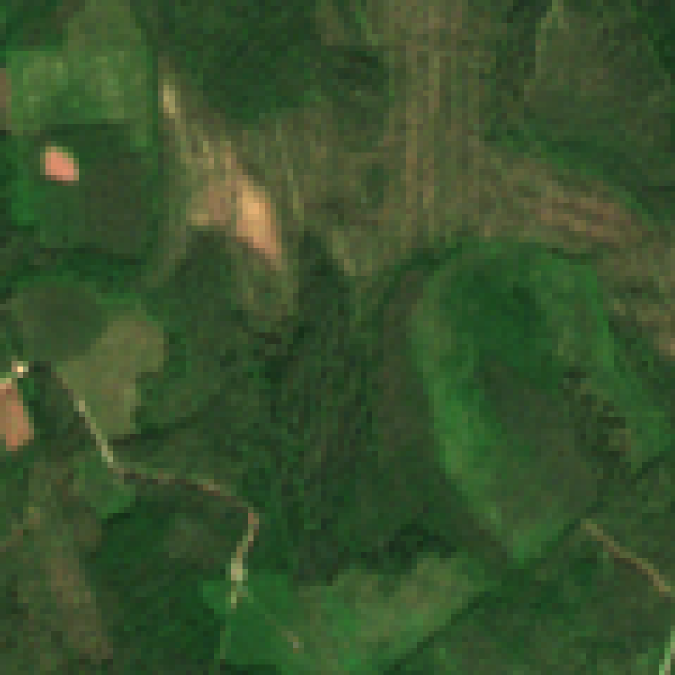}
        \caption{GSD: 10m per pixel}
        \label{fig:sf1_example_b}
    \end{subfigure}
    \caption{Example of variability induced by the GSD (SF1).}
    \label{fig:sf1_total}
\end{figure}
\begin{figure}[htbp]
    \centering
    \begin{subfigure}{0.49\linewidth}
        \centering
        \includegraphics[width=0.78\linewidth]{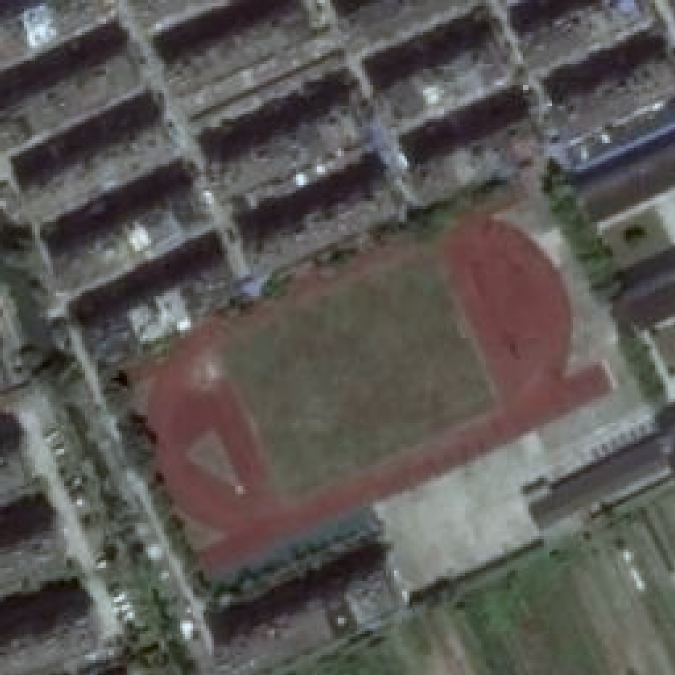}
        \caption{Land-focused scene.}
        \label{fig:sf2_example_a}
    \end{subfigure}
    \hfill
    \begin{subfigure}{0.49\linewidth}
        \centering
        \includegraphics[width=0.78\linewidth]{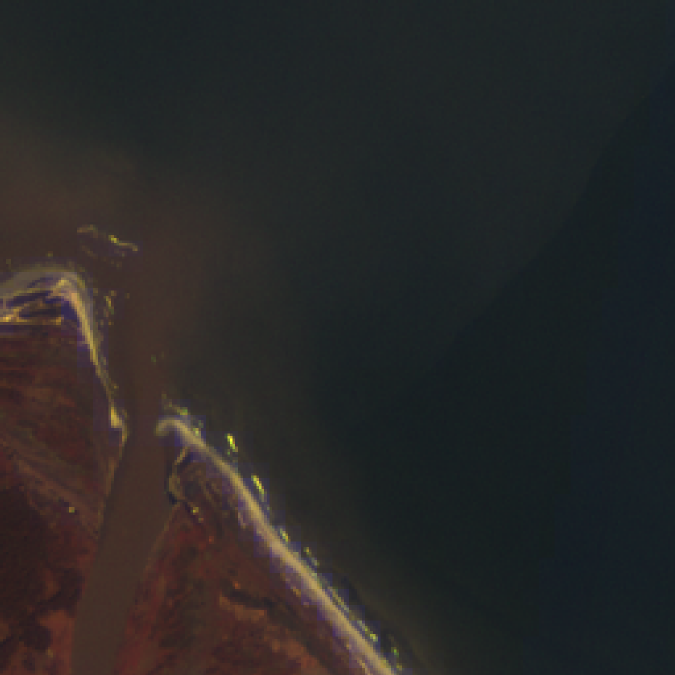}
        \caption{Marine-focused scene.}
        \label{fig:sf2_example_b}
    \end{subfigure}
    \caption{Example of the impact of the domain of interest (SF2).}
    \label{fig:sf2_total}
\end{figure}

\begin{figure}[htbp]
    \centering
    \begin{subfigure}{0.49\linewidth}
        \centering
        \includegraphics[width=0.78\linewidth]{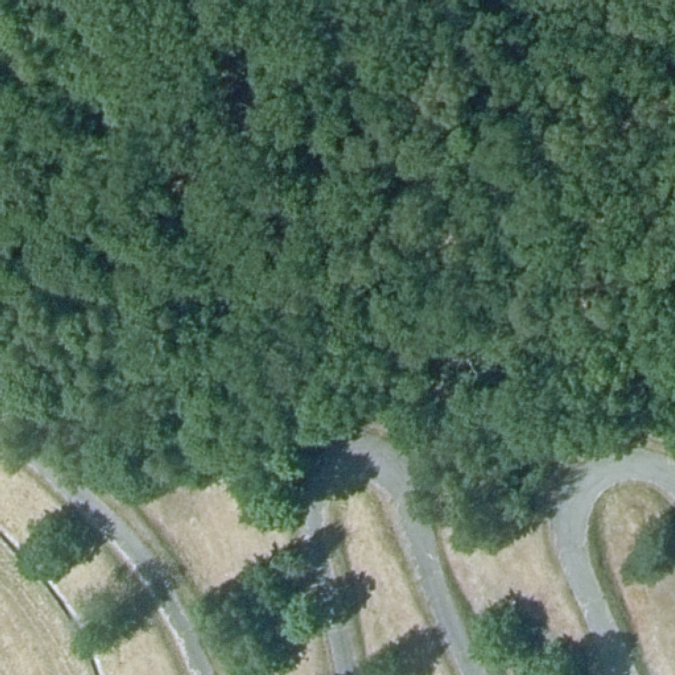}
        \caption{Target: Forest type.}
        \label{fig:sf3_example_a}
    \end{subfigure}
    \hfill
    \begin{subfigure}{0.49\linewidth}
        \centering
        \includegraphics[width=0.78\linewidth]{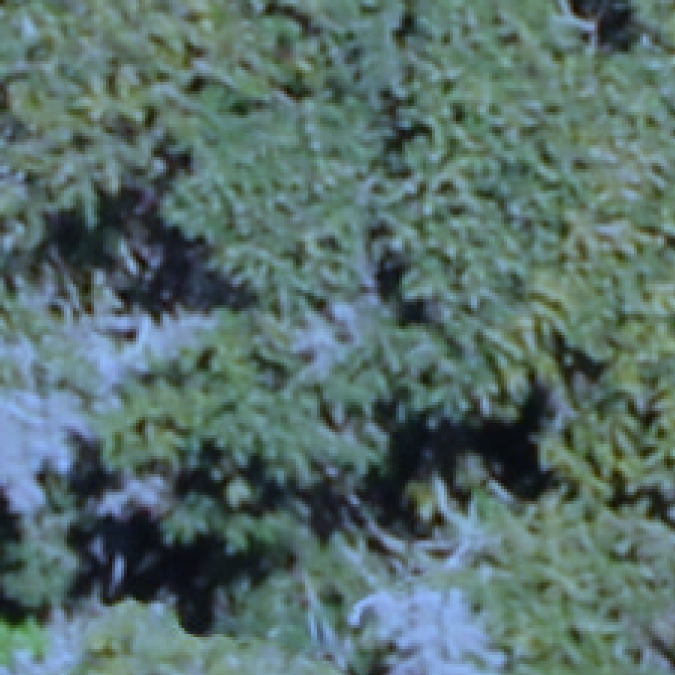}
        \caption{Target: Tree species.}
        \label{fig:sf3_example_b}
    \end{subfigure}
    \caption{Example of variability in target granularity (SF3).}
    \label{fig:sf3_total}
\end{figure}

\begin{figure}[htbp]
    \centering
    \begin{subfigure}{0.49\linewidth}
        \centering
        \includegraphics[width=0.82\linewidth]{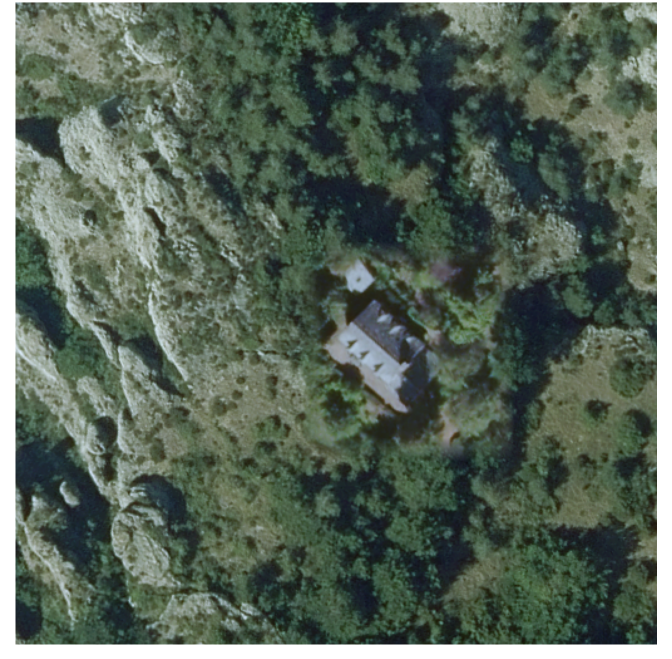}
        \caption{Mountainous forest area.}
        \label{fig:sf4_example}
    \end{subfigure}
    \hfill
    \begin{subfigure}{0.49\linewidth}
        \centering
        \includegraphics[width=0.82\linewidth]{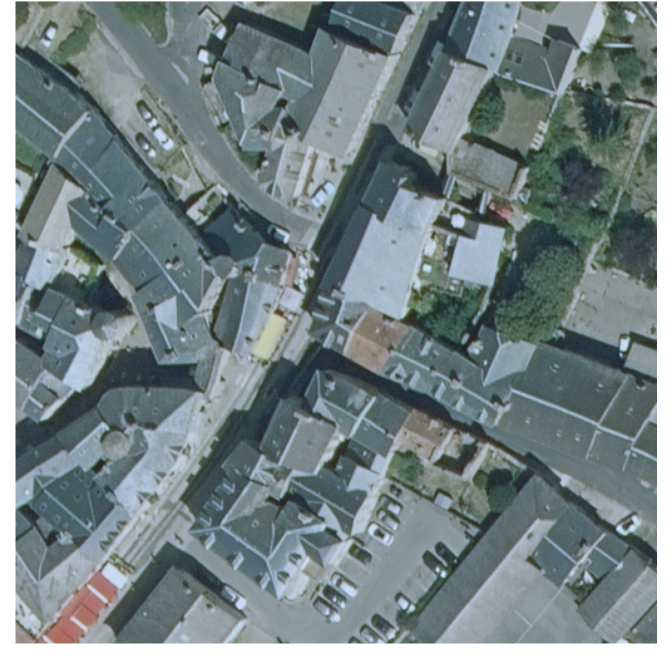}
        \caption{Urban area.}
        \label{fig:sf4_example_ok}
    \end{subfigure}
    \caption{Example of the impact of spatial arrangement (SF4).}
    \label{fig:sf4_total}
\end{figure}

In this section, we visually inspect the variability induced by the semantic factors defined in \cref{sub:semantic_factors}. 
\Cref{fig:sf1_total} highlights the variability induced by varying the GSD (SF1). 
\Cref{fig:sf1_example_a} presents a sample from the FLAIR dataset at 20cm per pixel, depicting a suburban area with fine detail, while \cref{fig:sf1_example_b} shows a sample from the BigEarthNet dataset at 10m per pixel, capturing coarser objects such as a forest.  \Cref{fig:sf2_total} showcases the impact of the domain of interest (SF2) in EO scenes, presenting a sample from MLRSNet focused on land cover (\cref{fig:sf2_example_a}), and a sample from the MARIDA dataset (\cref{fig:sf2_example_b}), which focuses on marine environments. Clearly, the objects existing in these two domains differ significantly. Similarly, \cref{fig:sf3_total} shows the impact of the target granularity, highlighting the visual difference of high-level targets like forests (\cref{fig:sf3_example_a}), which require less detail to identify, and low-level targets like tree species (\cref{fig:sf3_example_b}) that demand the highest possible level of detail.

Notably, as seen in the images, some SFs are highly interdependent; \eg, GSD (SF1) influences target granularity (SF3), as spatial resolution dictates the level of available detail.

\section{Uncertainty Representation Metrics}
\label{app:Β}
\textbf{\textit{LA@$1$} for multi-label classification:} Consider a sample $x$ and its nearest neighbor in the representation space denoted as $x^*$.
Let their class vectors be $\mathbf{c}, \mathbf{c^*} \in \{0, 1\}^K$, where $K$ is the number of classes. 
Each element in the class vectors indicates the presence ($1$) or absence ($0$) of a given class.
In the context of multi-label classification, we define $3$ metrics:

\noindent\textit{One-LA@$1$} assesses whether $x$ and $x^*$ share at least one common class and is defined as follows:
%$$
\begin{equation*}
\text{One-LA@$1$} = \mathbb{I}\left(\mathbf{c}^\top \mathbf{c^*} > 0\right),
\end{equation*}
%$$
where $\mathbb{I}$ is the indicator function, that equals $1$ if the condition is true, and $0$ otherwise.

\noindent\textit{All-LA@$1$} is a stricter criterion enforcing $x^*$ to contain all classes present in $x$. Formally:
%$$
\begin{equation*}    
\text{All-LA@$1$} = \mathbb{I}\left(\mathbf{c}^\top \mathbf{c^*} = \|\mathbf{c}\|_1\right),
\end{equation*}

%$$
where $\|\mathbf{c}\|_1$ denotes the sum of the elements in the class vector $\mathbf{c}$.

\noindent\textit{\%-LA@$1$} quantifies the proportion of classes of $x^*$ that match the classes of $x$:

\begin{equation*}
\text{\%-LA@$1$} = \frac{\mathbf{c}^\top \mathbf{c^*}}{\|\mathbf{c}\|_1}
\end{equation*}

This metric provides a balance between One-LA@$1$, which may be too lenient, and All-LA@$1$, which can be too strict.

\textbf{\textit{LA@$1$} for semantic segmentation:}
In the context of semantic segmentation, labels are represented as matrices 
$\mathbf{C}, \mathbf{C^*} \in \mathbb{R}^{m \times n}$, where $m$ and $n$ denote the height and width of the image respectively, and each element in the matrix corresponds to one of the $K$ possible classes. 
To capture different notions of semantic similarity in segmentation tasks, we introduce $4$ metrics:

\noindent\textit{All-LA@$1$} is a high-level metric, assessing whether the set of classes is shared between the images.
The class vectors $\mathbf{c}, \mathbf{c^*} \in \{0, 1\}^K$ are calculated, where each entry indicates the presence or absence of the respective class in the image.
The metric is calculated as in the multi-label scenario:
%$$
\begin{equation*}
\text{All-LA@$1$} = \mathbb{I}\left(\mathbf{c}^\top \mathbf{c^*} = \|\mathbf{c}\|_1\right)%.
\end{equation*}
%$$

\noindent\textit{Patches$_{p}$-LA@$1$} evaluates the spatial similarity between $x$ and $x^*$.
Each image is divided into a $p \times p$ grid of patches. 
For each patch, the majority class is identified, resulting in two matrices $\mathbf{M}, \mathbf{M^*} \in \mathbb{Z}^{P \times P}$.
Then, the metric computes whether corresponding patches in the two images share the same majority class:
%$$
\begin{equation*}
\text{Patches$_{p}$-LA@$1$} = \mathbb{I}\left(\mathbf{M_p} = \mathbf{M^*_p}\right)%, \\ 
\end{equation*}
%$$
This metric evaluates the spatial similarity between the images, by capturing the spatial alignment of their classes. 
In our study, we set $p=3$ to capture high-level context, but this can be adjusted depending on the task's requirements.

\noindent\textit{Probability Distance of LA@$1$ (PD-LA@1}) measures the contextual similarity between $x$ and $x^*$, by comparing the class distributions across the entire image.  
The distributions $\mathbf{p}, \mathbf{p^*} \in [0, 1]^K$ represent the percentages of each class in samples $x$ and $x^*$ respectively, and the metric is calculated as:
%$$
\begin{equation*}
\text{PD-LA@$1$}  = 1 - HD(\mathbf{p}, \mathbf{p^*}),
\end{equation*}
%$$ 
where HD is the Hellinger Distance
$HD(x,y) = \sqrt{\frac{1}{2} \sum_{i=1}^N \left( \sqrt{x_i} - \sqrt{y_i} \right)^2},$ which measures the similarity between two probability distributions.
This metric is used to measure the contextual similarity between the two images, regardless of the spatial position of the classes.

\noindent\textit{Patches$_{p}$ Probability Distance of LA@$1$ (Patches$_{p}$-PD-LA@1)} combines the spatial focus of Patches$_{p}$-LA@$1$ with the contextual similarity of PD-LA@$1$.
The image is divided into a $p \times p$ grid of patches and for each patch $i, j \in \{1, \ldots, p\}$, the class distributions $\mathbf{p}_{ij}$ and $\mathbf{p^*}_{ij}$  are calculated.
The $HD_{ij}$ is then calculated for each $i, j$, and the final metric is the average of these distances across all patches:
%$$
\begin{equation*}
\text{Patches$_{p}$-PD-LA@$1$} = 1 - \frac{1}{P^2} \sum_{i=1}^P \sum_{j=1}^P HD_{ij}.
\end{equation*}
%$$
This metric captures both the spatial and contextual similarities between the two images, offering a comprehensive view of how well the two images align, both in terms of pixel position and overall structure.

The LA@$1$ for a dataset is calculated as the mean LA@$1$ across its samples.
Binary LA@$1$ metrics (One-LA@$1$, All-LA@$1$, Patches$_{p}$-LA@$1$) quantify the percentage of representations whose nearest neighbor is semantically similar.
In contrast, PD-LA@$1$ and Patches$_{p}$-PD-LA@$1$ represent mean probability distances, while \%-LA@$1$ reflects mean proportions.
This emphasizes that the values of the metrics are not directly comparable, as they measure different elements.

\section{Coefficient of Predictive Ability (CPA)}
\label{app:cpa}
Traditional ROC analysis is designed for binary classification tasks.
The Universal ROC (UROC) curves and the associated Coefficient of Predictive Ability (CPA) extend this framework to any linearly ordered outcome, including binary, ordinal, mixed discrete-continuous, and continuous variables, thereby generalizing ROC analysis \cite{gneiting_receiver_2021}.

Generalizing the binary setting, the problem is transformed into a sequence of binary classification tasks.
Given bivariate data \((x_i, y_i)\) for \(i = 1, \dots, n\), where \(x_i\) represents a predictor and \(y_i\) a continuous outcome, \(m\) unique values of $y_i$ \(z_1 < \dots < z_m\) with $m<n$ are defined.

To construct the UROC framework, the real-valued outcomes are converted into binary indicators $\mathbbm{1}\{y_1\geq\theta\},\dots\mathbbm{1}\{y_n\geq\theta\}$ for threshold values \(\theta \in \{z_2, \dots, z_m\}\). 
This results in \(m-1\) derived binary classification problems of the form 

\[
(x_i, 1\{y_i \geq z_{c+1}\}), \quad c = 1, \dots, m-1.
\]

Each of these $m-1$ binary classification problems admits a standard ROC curve, which can be sequentially visualized as a "ROC movie." 
To summarize these %multiple
ROC curves into a single representation, the UROC curve is defined as a weighted average of the individual ROC curves, providing a unified representation of predictive performance across continuous outcomes.

The CPA, defined as the area under the UROC curve, serves as a generalization of AUROC for continuous outcomes.
Particularly, CPA is a weighted average of the AUROC values for the derived binary problems in the very same way that the UROC curve is a weighted average of the classical ROC curves that constitute the ROC movie.
Notably, in the case of strictly binary outcomes, CPA reduces to AUROC, preserving interpretability within the classical ROC framework.

For further details, we refer the reader to the original work by Gneiting et al. \cite{gneiting_receiver_2021}.

\begin{table*}\centering
\scriptsize
\begin{tabular}{c c c c c c}\toprule
Investigation Target & Section & Figure & Model & Pretraining Datasets & Inference Datasets\\
\midrule
Impact of SF1 & \cref{sub:concept} & \cref{fig:resolution1} & ViT-Large & BigEarthNet & All \\
\midrule
Impact of SF1 & \cref{sub:concept} & \cref{fig:resolution2} & ViT-Large & BigEarthNet & Flair \\
\midrule
Impact of SF2 & \cref{sub:results} & \cref{fig:results} & All & ImageNet, BigEarthNet, Flair & MARIDA \\
\midrule
Impact of SF3 & \cref{sub:concept} & \cref{fig:results} & All & BigEarthNet-5 & All \\
\midrule
Impact of SF4 & \cref{sub:concept} & \cref{fig:results} & All & BigEarthNet-5 & All \\
\midrule
Usefulness of uncertainties & \cref{sub:downstream} & \cref{fig:discard-test} & ViT-Large & ImageNet, BigEarthNet, Flair & All \\
\midrule
Usefulness of uncertainties & \cref{sub:downstream} & \cref{fig:discard_flair} & ViT-Large & ImageNet, BigEarthNet, Flair & Flair \\
\midrule
Localized Uncertainty & \cref{sub:per_patch_unc} & \cref{fig:per_pixel} & ViT-Large & BigEarthNet & Flair, MARIDA \\
\midrule
Noisy Data & \cref{sub:noisy} & \cref{fig:noisy_data_eval} & All & ImageNet, Flair & BigEarthNet, noisy BigEarthNet  \\
\bottomrule
\end{tabular}
\caption{Overview of the experiments conducted in this study. Each investigation target is accompanied by references to the relevant sections, figures, models, and the corresponding pre-training and inference datasets used.}
\label{tab:link_semantic_factors}
\end{table*}

\begin{table*}[]
    \centering%\fontsize{9}{}\selectfont
    \scalebox{0.8}{ 
    \begin{tabular}{ccccccc}
        \hline
        Figure/Table & Pretraining Dataset & Input Res. & Model & unc$\_$width & weight decay & modality\\ \hline
        \multirow{4}{*}{\cref{fig:results}} & \multirow{4}{*}{BigEarthNet} & \multirow{4}{*}{120} & ViT-Tiny & 512 & 0.1 & RGB\\
        & & & ViT-Small & 512 & 0.1 & RGB\\
        & & & ViT-Base & 512 & 0.1 & RGB\\
        & & & ViT-Large & 512 & 0.1 & RGB\\ \hline
        \multirow{4}{*}{\cref{tab:app_bigearthnet}} & \multirow{4}{*}{BigEarthNet} & \multirow{4}{*}{120} & ViT-Tiny & 512 & 0.1 & SAR\\
        & & & ViT-Small & 512 & 0.1 & SAR\\
        & & & ViT-Base & 512 & 0.1 & SAR\\
        & & & ViT-Large & 512 & 0.1 & SAR\\ \hline
        \multirow{4}{*}{\cref{tab:app_bigearthnet}} & \multirow{4}{*}{BigEarthNet} & \multirow{4}{*}{120} & ViT-Tiny & 512 & 0.1 & MS\\
        & & & ViT-Small & 512 & 0.1 & MS\\
        & & & ViT-Base & 512 & 0.1 & MS\\
        & & & ViT-Large & 512 & 0.1 & MS\\ \hline
        \multirow{4}{*}{\cref{fig:results}} & \multirow{4}{*}{BigEarthNet-5} & \multirow{4}{*}{120} & ViT-Tiny & 256 & 0.01 & RGB \\
        & & & ViT-Small & 512 & 0.1 & RGB\\
        & & & ViT-Base & 512 & 0.01 & RGB\\
        & & & ViT-Large & 512 & 0.1 & RGB\\ \hline
        \multirow{4}{*}{\cref{fig:results}} & \multirow{4}{*}{Flair} & \multirow{4}{*}{120} & ViT-Tiny & 256 & 0.5 & RGB\\
        & & & ViT-Small & 256 & 0.5 & RGB\\
        & & & ViT-Base & 256 & 0.5 & RGB\\
        & & & ViT-Large & 256 & 0.5 & RGB\\ \hline
        \cref{fig:resolution1} & BigEarthNet & 60 & ViT-Large & 512 & 0.5 & RGB\\
        %\hline
        \cref{fig:resolution1} & BigEarthNet & 30 & ViT-Large & 512 & 0.01 & RGB\\
        %\hline
       \cref{fig:resolution1} & BigEarthNet & 16 & ViT-Large & 512 & 0.1 & RGB\\
    \end{tabular}
    }
\caption{Model configurations and settings used for training the uncertainty modules. The "Figure/Table" column indicates the corresponding figure or table in the main text or supplementary material associated with each experiment.}    \label{tab:hyper}
\end{table*}

\begin{table*}[!htp]\centering
\scriptsize
\begin{tabular}{l l rr rr rr rr}\toprule
\multirow{3}{*}{Metric} & \multirow{3}{*}{Modality} & \multicolumn{2}{c}{ViT - Tiny} & \multicolumn{2}{c}{ViT - Small} & \multicolumn{2}{c}{ViT - Base} & \multicolumn{2}{c}{ViT - Large} \\
& & LA@1 & LA-CPA & LA@1 & LA-CPA & LA@1 & LA-CPA & LA@1 & LA-CPA \\
\midrule
\multirow{3}{*}{One} 
    & RGB & 0.995 & 0.429 & 0.996 & 0.437 & 0.997 & 0.427 & 0.998 & 0.544 \\
    & SAR & 0.975 & 0.435 & 0.978 & 0.427 & 0.979 & 0.403 & 0.981 & 0.418 \\
    & MS  & 0.996 & 0.402 & 0.997 & 0.366 & 0.997 & 0.431 & 0.997 & 0.353 \\
\midrule
\multirow{3}{*}{All} 
    & RGB & 0.526 & 0.657 & 0.592 & 0.646 & 0.664 & 0.616 & 0.716 & 0.607 \\
    & SAR & 0.397 & 0.665 & 0.42  & 0.659 & 0.463 & 0.641 & 0.463 & 0.641 \\
    & MS  & 0.558 & 0.650 & 0.63  & 0.632 & 0.609 & 0.613 & 0.723 & 0.615 \\
\midrule
\multirow{3}{*}{\%} 
    & RGB & 0.796 & 0.607 & 0.827 & 0.596 & 0.857 & 0.569 & 0.88  & 0.575 \\
    & SAR & 0.714 & 0.600 & 0.727 & 0.595 & 0.748 & 0.582 & 0.748 & 0.582 \\
    & MS  & 0.815 & 0.600 & 0.845 & 0.583 & 0.834 & 0.571 & 0.885 & 0.562 \\
\bottomrule
\end{tabular}
\caption{LA@$1$ and LA-CPA for Multispectral (MS), Synthetic Aperture (SAR) and RGB data for BigEarthNet pretraining and inference.}
\label{tab:app_bigearthnet}
\end{table*}

\section{Discard Test}
\label{app:discard_test}

The Discard Test is a diagnostic tool used to assess the quality of a model’s uncertainty estimates by iteratively removing the most uncertain predictions from a test set and measuring the resulting change in model error. 
The fundamental principle behind this test is that if a model’s uncertainty estimates are reliable, the most uncertain predictions should correspond to higher errors, thus removing them should lead to an improvement in overall model performance.
The exact steps of the test are the following: 
\begin{enumerate}
    \item Model predictions are ranked in descending order based on their associated uncertainty estimates.
    \item The ranked samples are divided into equal-sized batches according to a predefined discard fraction.
    \item The most uncertain batch is removed from the set.
    \item The model’s error is recalculated on the remaining test samples.
    \item Steps 3–4 are repeated iteratively until all samples have been discarded.
\end{enumerate}

This process generates a curve that visualizes how the model error changes as more uncertain predictions are excluded. 
An effective uncertainty estimation method should result in a monotonically decreasing error curve, indicating that the most uncertain samples also tend to have higher errors. 
Deviations from these trends, such as non-monotonic error curves, suggest that the uncertainty estimates are not fully reliable, as removing uncertain predictions does not consistently enhance model performance.
In this study, we use $200$ discard fractions, so the steps are repeated $200$ times for each dataset and pre-trained model.
Moreover, we use the model loss as a measure of error.

\begin{figure*}
    \centering
    \includegraphics[width=1.0\linewidth]{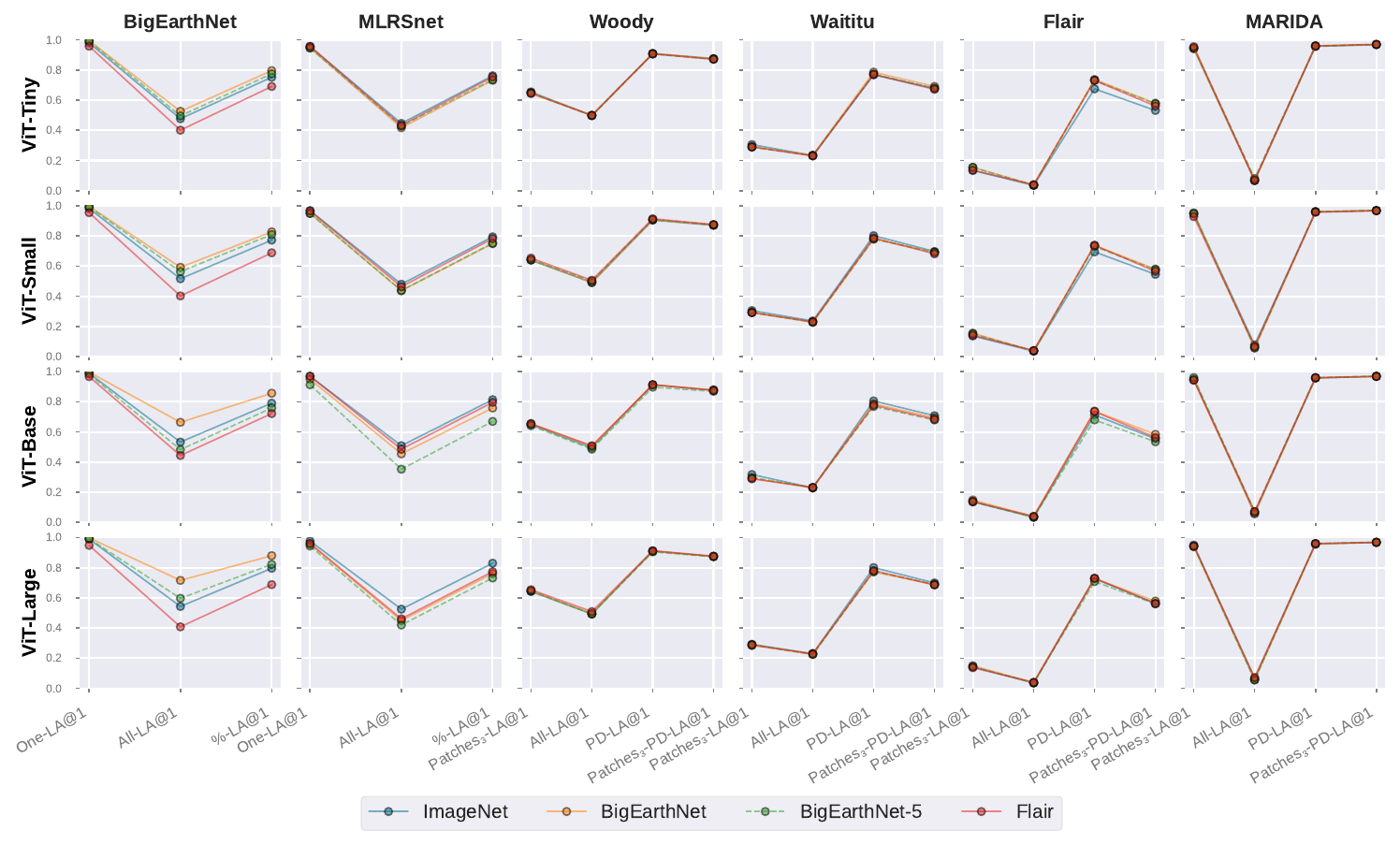}
    \caption{Performance evaluation of LA@$1$ for all ViT variants and inference datasets across all LA metrics, with higher values representing better performance. Colors indicate different pre-training datasets. EO-pretrained models and ImageNet pre-trained models show comparable behavior.}
    \label{fig:results_la@1}
\end{figure*}

\begin{table*}[!htp]\centering
\scriptsize
\begin{tabular}{l l cc cc cc cc}\toprule
\multirow{2}{*}{Dataset} & \multirow{2}{*}{Modality} & \multicolumn{2}{c}{ViT - Tiny} & \multicolumn{2}{c}{ViT - Small} & \multicolumn{2}{c}{ViT - Base} & \multicolumn{2}{c}{ViT - Large} \\
& & micro F1 (\%) & macro F1(\%) & micro F1 (\%) & macro F1(\%)  & micro F1 (\%) & macro F1(\%)  &micro F1 (\%) & macro F1(\%)  \\
\midrule
\multirow{3}{*}{BigEarthNet} 
    & RGB & 73.25 & 59.65 & 73.78 & 61.94 & 73.19 & 60.11 & 73.32 & 60.81\\
    & SAR & 68.48 &54.76 & 68.33 & 55.92 & 68.30 & 55.80 & 68.94 & 55.28\\
    & MS  & 74.32 & 62.13 & 73.92 & 62.36 & 74.20 & 62.28 & 74.22 & 62.36 \\
\midrule
BigEarthNet-5 & RGB & 84.19 & 60.67 & 84.73 & 81.61 &84.21 &70.93 & 84.10 & 81.07\\
\midrule
Flair & RGB &  85.21 & 57.62 & 85.87 & 59.54 & 85.20 & 59.23 & 85.78 & 58.97\\
\midrule
MLRSNet & RGB & 91.50 & 92.37& 91.91 & 98.45 & 91.45 & 98.35 & 98.46  & 91.94 \\
\bottomrule
\end{tabular}
\caption{Performance of supervised models whose representations were used for training the uncertainty modules that were used for creating \cref{fig:results} of the main text and \cref{fig:results_la@1} of SM.}
\label{tab:pretrained_supervised_models}
\end{table*}

The results are accompanied by the provision of Monotonicity Fraction (MF).
MF measures how often model performance improves as more uncertain samples are discarded and is computed as:
\begin{equation*}
MF = \frac{1}{N_{f}-1}\sum_{i=1}^{N_{f}-1}\mathbbm{1}(\epsilon_{i}\geq\epsilon_{i+1}),
\end{equation*}%$$
where $\mathbbm{1}$ is the indicator function, and $\epsilon_i$ is the model error (here the loss) at discard fraction $i$.
$N_f$ denotes the total number of discard fractions considered.
An MF value of $1$ indicates perfect monotonicity.
An ideal uncertainty estimation method would yield a high MF (indicating consistent performance improvement). 

\begin{table}\centering
\scriptsize
\begin{tabular}{c c c c}\toprule
Input Res. & Modality & micro F1 (\%) & macro F1(\%) \\
\midrule
120 &  RGB & 73.78 & 61.94\\
\midrule
60 &  RGB & 70.54 &  57.74 \\
\midrule
30 & RGB &66.00 & 52.06 \\
\midrule
16 & RGB &60.26 & 44.58 \\
\bottomrule
\end{tabular}
\caption{Performance of supervised models whose representations used for training the uncertainty models under varying GSD. These models refer to \cref{fig:resolution1} of the main text.}
\label{tab:input_res_inv}
\end{table}

\begin{table}\centering
\scriptsize
\begin{tabular}{c c cc }\toprule
Dataset & {Modality} & micro F1 (\%) & meanIOU (\%)\\
\midrule
Flair & RGB & 72.59 & 56.98 \\
\midrule
Marida & RGB & 99.17 & 98.35 \\
\midrule
Waititu & RGB & 84.20 & 72.72 \\
\midrule
Woody & RGB & 93.45 & 87.71 \\
\bottomrule
\end{tabular}
\caption{Performance of U-Net, with a ResNet-50 backbone, trained via supervised learning for semantic segmentation tasks. The alignment of zero-shot uncertainties was assessed with the losses of these models. They refer to \cref{fig:discard-test} of the main text.}
\label{tab:pretrained_segmentation_models}
\end{table}

\section{Overview of the experimental design}
\label{app:semantic_factors}

\Cref{tab:link_semantic_factors} summarizes the experimental design employed to evaluate the generalization of representation uncertainty. The table offers detailed references to the relevant investigation targets and the sections, figures, models, and dataset configurations that support each case, thereby facilitating traceability of the experiments.

\section{Uncertainty Module training details}
\label{app:unc_module}
The uncertainty module was implemented as an MLP with two hidden layers, applied on top of the learned representations. 
Each linear layer is followed by a LeakyReLU activation, and the final layer uses a Softplus activation to ensure the uncertainties remain positive, as described in the original paper.
The uncertainty width $unc\_width$, \ie the width of the linear layers, is set to either $256$ or $512$, tuned individually for each model. %, with possible values of $256$ or $512$.
A full overview of the hyperparameters used across all experiments can be found in \cref{tab:hyper}.

Our models were trained for $1000$ epochs in each configuration.
The learning rate was warmed up with a constant value of $0.0001$ for $50$ epochs and then decayed to $1e-8$ for the remaining epochs.
Weight decay was tuned separately for each model.
AdamW was used as the optimizer with $\beta_1=0.8$ and $\beta_2=0.95$.
No augmentations were applied during the training of the uncertainty module.

\section{Multispectral \& Synthetic Aperture Radar Data Pretraining}
\label{app:ms_sar}

In the main text, we evaluated our framework's results using only the RGB channels to ensure a fair comparison with models pretrained on RGB optical data and to accommodate datasets with diverse spectral characteristics extending beyond RGB. 
However, recognizing the importance of applications beyond the RGB spectral bands, we also pretrain models on MS and SAR data and publicly release the pretrained weights.
These models are pretrained on BigEarthNet, a dataset with both MS and SAR modalities, and their performance is evaluated on the same dataset, facilitating the comparison with the RGB training setup.
In \cref{tab:app_bigearthnet}, we report the results of this evaluation.
While LA@$1$ is consistently better for MS and RGB modalities compared to SAR, the LA-CPA is slightly better in SAR modalities, especially in larger models (ViT-Base, ViT-Large).
This is a preliminary indication that our framework can effectively extend to other setups, yet further examination on additional datasets and setups is necessary to test the uncertainty generalization and draw a more definitive conclusion.

\section{LA@1 across datasets}
\label{app:recall@1}

\Cref{fig:results_la@1} summarizes the LA@$1$ results across different metrics, similar to \cref{fig:results}, which presents the LA-CPA results discussed in the main text. 
As highlighted in the main text, LA@$1$ remains consistent across models irrespective of the pre-training dataset, with ImageNet feature extractors producing robust representations that yield LA@$1$ values comparable to those trained on EO datasets.

\section{Visualization of Samples with High/Low uncertainty}
\label{app:unc_samples}

\begin{figure*}[ht]
    \centering
    \includegraphics[width=0.9\linewidth]{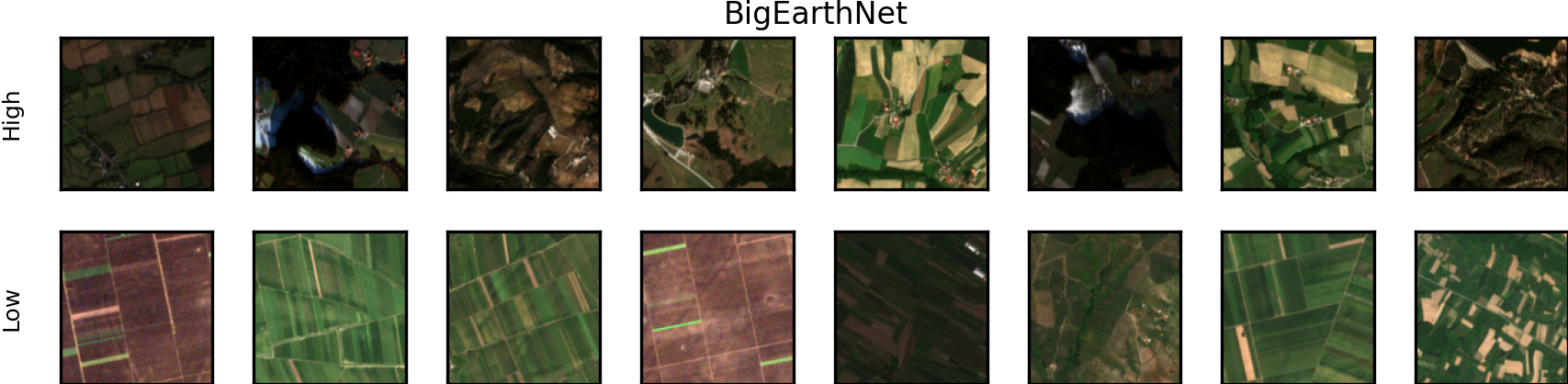} \\
    \vspace{0.15cm} % Adds space between images
    \includegraphics[width=0.9\linewidth]{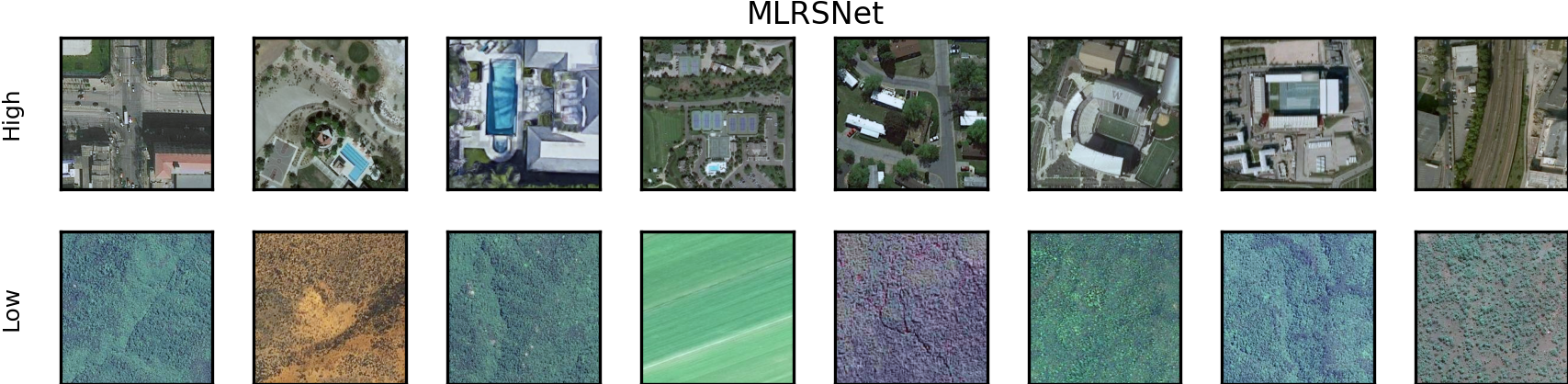} \\
    \vspace{0.15cm} % Adds space between images
    \includegraphics[width=0.9\linewidth]{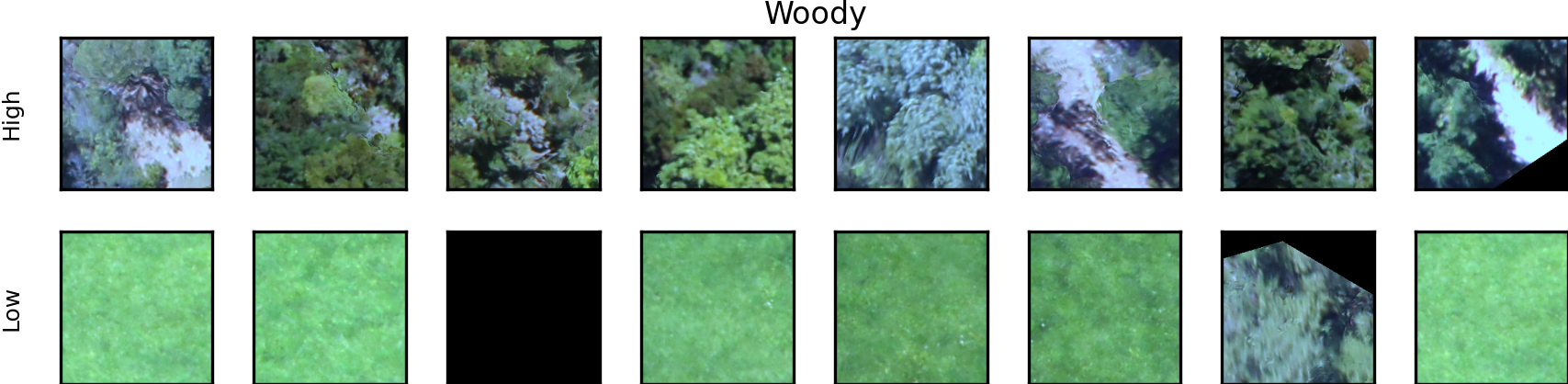} \\
    \vspace{0.15cm} % Adds space between images
    \includegraphics[width=0.9\linewidth]{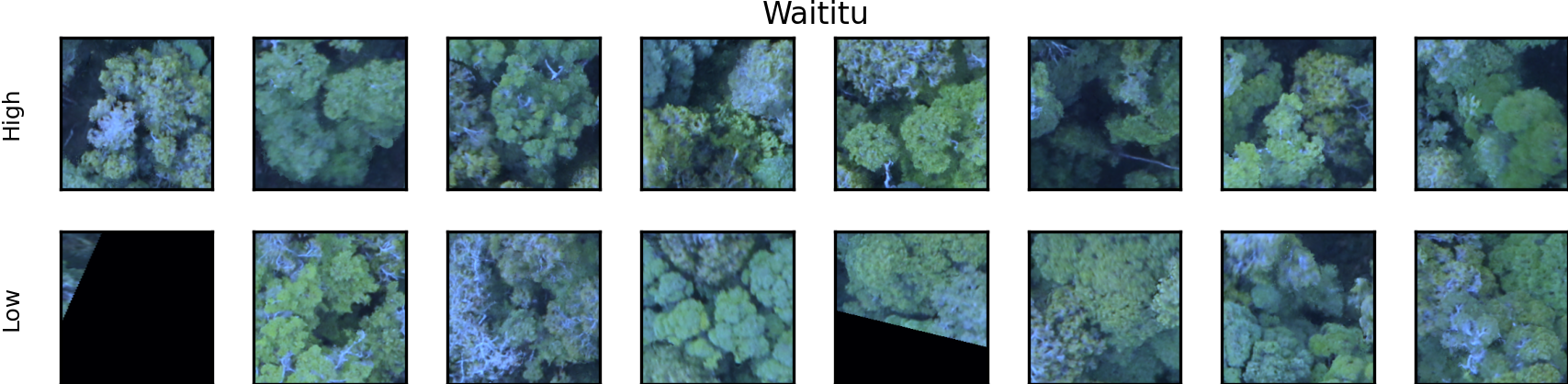} \\
    \vspace{0.15cm} % Adds space between images
    \includegraphics[width=0.9\linewidth]{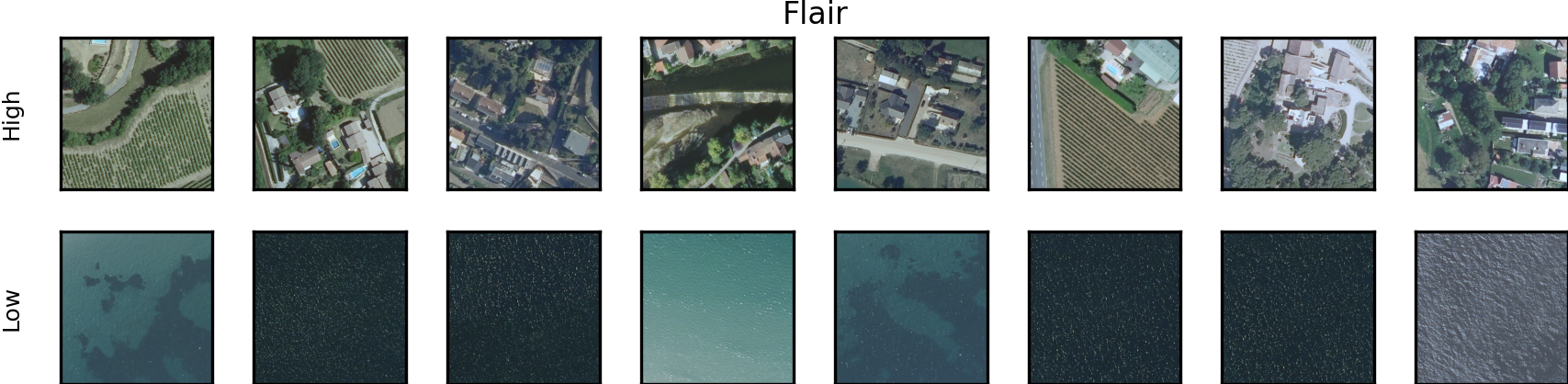} \\
    \caption{Samples with high/low downstream representation uncertainty across datasets used in this study. The uncertainty estimates were extracted from a ViT-Large pretrained on Flair.}
    \label{fig:unc_samples_sm}
\end{figure*}

\Cref{fig:unc_samples_sm} presents samples with high/low uncertainties across all datasets, as estimated by ViT-Large pretrained on Flair, providing a qualitative perspective on the performance of pretrained uncertainties. 
%Uncertainty for these samples was calculated from a Flair pretrained model.

\section{Pretrained supervised models}
\label{app:pre_trained}

In this section, we provide an overview of the performance of the supervised models used as backbone networks in our study. 
These models were used to extract the representations for training the uncertainty modules and to create the discard test plots.
In \cref{tab:pretrained_supervised_models}, we summarize the results for the pretrained models used for creating \cref{fig:results} of the main text and \cref{fig:results_la@1} in the SM.
The performance of the models used to investigate the impact of GSD on generalization, shown in \cref{fig:resolution1} and \ref{fig:resolution2} of the main text, is presented in \cref{tab:input_res_inv}. 
Finally, the results of the models used to examine the reliability of zero-shot uncertainties in downstream tasks, as shown in \cref{fig:discard-test}, are detailed in \cref{tab:pretrained_segmentation_models}.

\section{Localized Uncertainty Samples}
\label{app:localized}

\begin{figure*}
    \centering
    \includegraphics[width=0.8\linewidth]{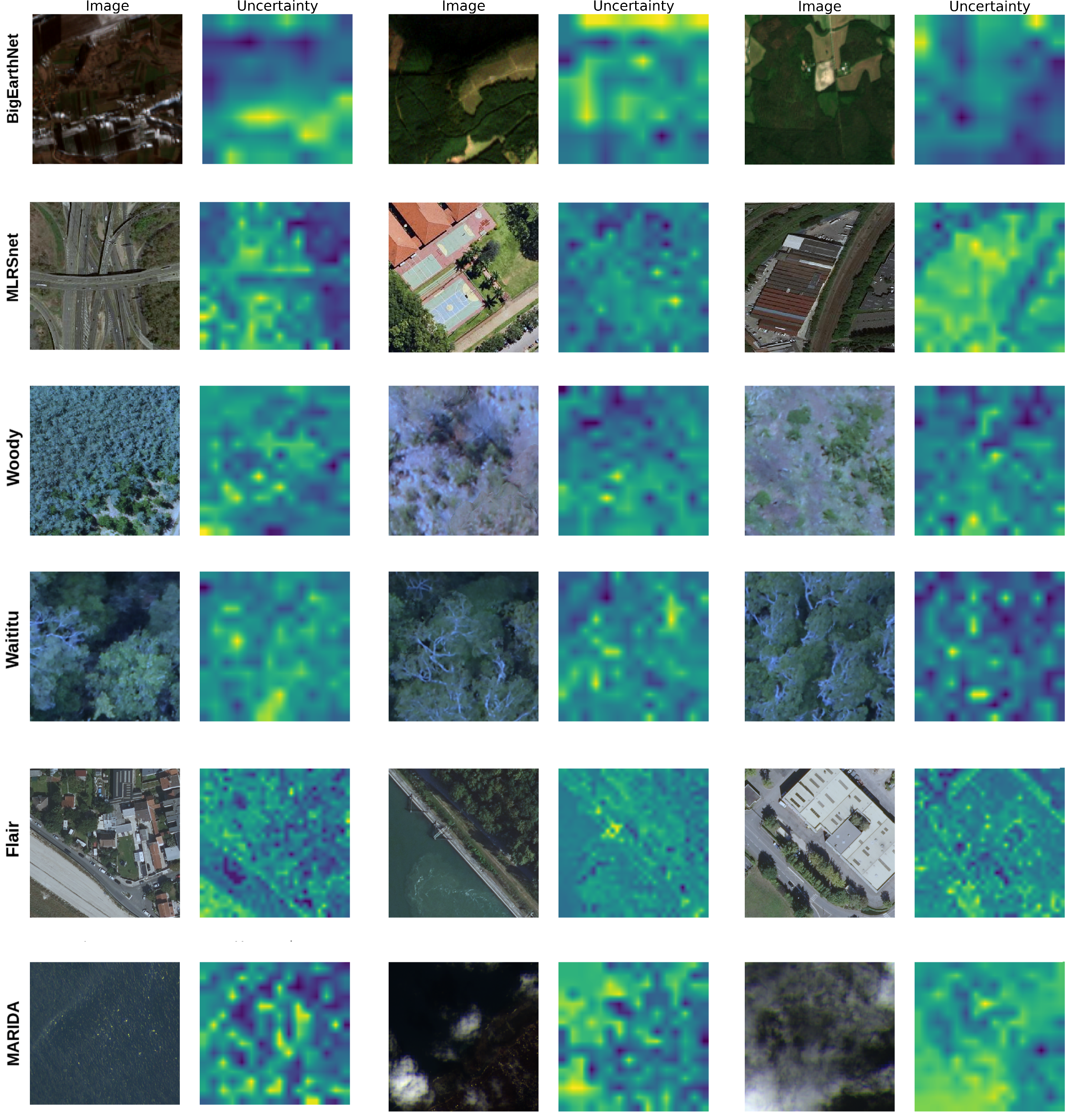}
    \caption{Example samples along with their zero-shot localized uncertainty estimates, as extracted by a ViT-Large pretrained on BigEarthNet.}
    \label{fig:loc_unc_sm}
\end{figure*}

In \cref{fig:loc_unc_sm} we provide uncertainty samples together with their localized uncertainty estimates, as extracted from a ViT-Large pretrained on BigEarthNet.

\section{Uncertainty distribution between upstream and downstream tasks}
\label{sec:signs_aleatoric}
\begin{figure}
  \includegraphics[width=0.95\linewidth]{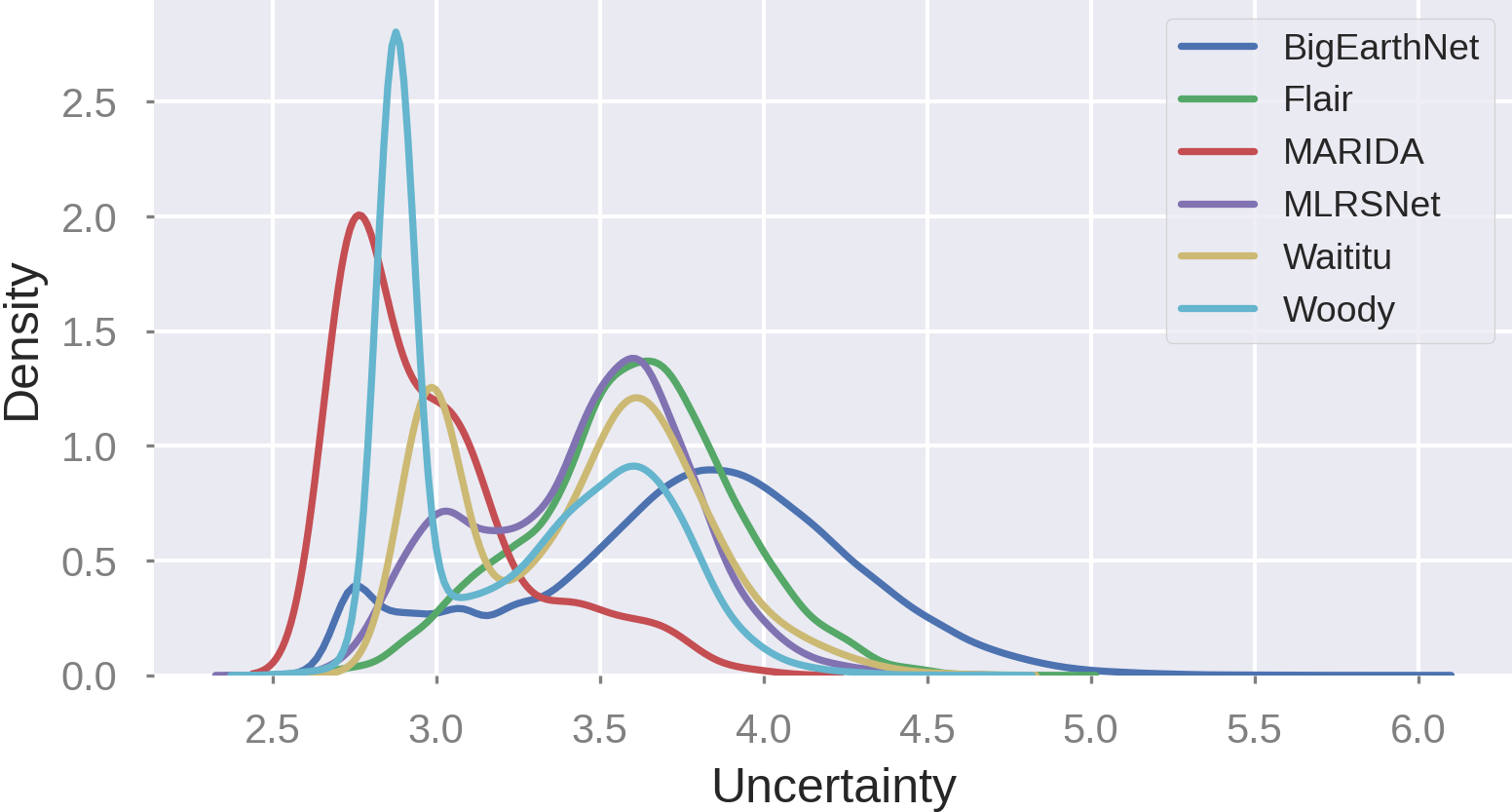}
  \caption{Densities of pretrained uncertainties, generated from a ViT-Large pretrained on BigEarthNet.}
  \label{fig:density}
\end{figure}

In Sec. 4.5 of \cite{kirchhof2024pretrained} the authors showed that predicted uncertainties capture aleatoric, and not epistemic uncertainty.
The presence of aleatoric uncertainty is validated in our experiments by \cref{fig:noisy_data_eval}, where the noisy subset of BigEarthNet exhibits higher uncertainty than its noise-free counterpart, despite coming from the same dataset.
To further solidify %reinforce
the lack of epistemic signals, we replicate the analysis from \cite{kirchhof2024pretrained} using ViT-Large pretrained on BigEarthNet and compare its uncertainty distribution with the ones in downstream tasks (Fig.~\ref{fig:density}). Despite the distribution shift, uncertainties on BigEarthNet span a wide range and are higher than those on downstream datasets, refuting the assumption that they reflect epistemic uncertainty. Notably, MARIDA exhibits the lowest uncertainty while coming from a very distinct EO domain.

\end{document}